\documentclass[lettersize,journal]{IEEEtran}

\usepackage{cite} 
\usepackage{amsmath,amssymb,amsfonts}
\usepackage{adjustbox} 
\usepackage[ruled,vlined,linesnumbered]{algorithm2e}
\usepackage{graphicx}
\usepackage[caption=false]{subfig}
\usepackage{booktabs}
\usepackage{bbm}
\usepackage{multirow}
\usepackage{caption}
\usepackage{float}
\usepackage{tikz}
\usetikzlibrary{arrows.meta,positioning,shapes.geometric,calc,backgrounds,fit}
\definecolor{steelblue}{RGB}{70,130,180}
\definecolor{coral}{RGB}{232,115,90}
\definecolor{sagegreen}{RGB}{123,174,127}
\definecolor{warmgray}{RGB}{160,160,160}
\definecolor{softpurple}{RGB}{155,142,193}
\tikzset{
  twm/box/.style={draw=black, thin, rounded corners=2pt, minimum height=5mm,
    minimum width=22mm, font=\scriptsize, text=black, align=center, inner sep=2pt},
  twm/annot/.style={font=\scriptsize\itshape, text=black!55, align=center},
  twm/arr/.style={-{Stealth[length=2.5pt]}, thin},
  twm/darr/.style={-{Stealth[length=2.5pt]}, thin, dashed, black!50},
}
\usepackage[hidelinks]{hyperref}
\usepackage{url}
\usepackage{mdframed}
\usepackage{tabularx}
\usepackage{makecell}
\hypersetup{
  colorlinks=false,          
  citebordercolor=green,      
  linkbordercolor=red,     
  pdfborder={0 0 1}          
}
\usepackage{array}
\usepackage{pifont}
\newcommand{\cmark}{\ding{51}}  
\newcommand{\pmark}{\ding{109}} 
\newcommand{\xmark}{\textemdash} 

\title{Telecom World Models: Unifying Digital Twins, Foundation Models, and Predictive Planning for 6G}

\author{
Hang~Zou,
Yuzhi~Yang,
Lina~Bariah,
Yu~Tian,
Yuhuan~Lu,
Bohao~Wang,
Anis~Bara,
Brahim~Mefgouda,
Hao~Liu,
Yiwei~Tao,
Sergy~Petrov,
Salma~Cheour,
Nassim~Sehad,
Sumudu~Samarakoon,
Chongwen~Huang,
Samson~Lasaulce,
Mehdi~Bennis,
and~M\'erouane~Debbah

\thanks{H. Zou, Y. Yang, L. Bariah, Y. Tian,  Y. Lu, A. Bara, B. Mefgouda, H. Liu, Y. Tao,  S. Petrov, S. Cheour and M. Debbah are with Research Institute for Digital Future, Khalifa University, 127788 Abu Dhabi, UAE (e-mails: \{firstname.lastname\}@ku.ac.ae).
}
\thanks{B. Wang and C. Huang are with College of Information Science and Electronic Engineering, Zhejiang University, 310027, Hangzhou, China (email: \{bohaowang, chongwen.huang\}@zju.edu.cn)}
\thanks{N. Sehad is with University of Oulu and Aalto University (email:\{nassim.sehad\}@oulu.fi)}
\thanks{S. Lasaulce is with Universit\'{e} de Lorraine, CNRS, CRAN, F-54000 Nancy, France (email: samson.lasaulce@univ-lorraine.fr).}
\thanks{S. Samarakoon and M. Bennis are with University of Oulu, Finland (email: sumudu.samarakoon, mehdi.bennis@oulu.fi).}}
  
\begin{document}

\maketitle

\begin{abstract}
The integration of machine learning tools into telecom networks, has led to two prevailing paradigms, namely, language-based systems, such as Large Language Models (LLMs), and physics-based systems, such as Digital Twins (DTs). While LLM-based approaches enable flexible interaction and automation, they lack explicit representations of network dynamics. DTs, in contrast, offer a high-fidelity network simulation, but remain scenario-specific and are not designed for learning or decision-making under uncertainty. This gap becomes critical for 6G systems, where decisions must take into account the evolving network states, uncertainty, and the cascading effects of control actions across multiple layers. In this article, we introduce the {Telecom World Model}~(TWM) concept, an architecture for learned, action-conditioned, uncertainty-aware modeling of telecom system dynamics. We decompose the problem into two interacting worlds, a controllable system world consisting of operator-configurable settings and an external world that captures propagation, mobility, traffic, and failures. We propose a three-layer architecture, comprising a field world model for spatial environment prediction, a control/dynamics world model for action-conditioned Key Performance Indicator (KPI) trajectory prediction, and a telecom foundation model layer for intent translation and orchestration. We showcase a comparative analysis between existing paradigms, which demonstrates that TWM jointly provides telecom state grounding, fast action-conditioned roll-outs, calibrated uncertainty, multi-timescale dynamics, model-based planning, and LLM-integrated guardrails. Furthermore, we present a proof-of-concept on network slicing to validate the proposed architecture, showing that the full three-layer pipeline outperforms single-world baselines and accurately predicts KPI trajectories.
\end{abstract}

\begin{IEEEkeywords}
Telecom world models, world models, digital twins, LLMs, generative agents.
\end{IEEEkeywords}

\section{Introduction}
Recent advances in large language models (LLMs) and LLM-based agents have significantly expanded the role of machine learning (ML) in engineering systems, enabling natural language interactions, code generation, and automated workflows \cite{brown2020gpt3, comanici2025gemini, chatnet2023}. In the telecommunications community, several frameworks studied the integration of ``TelecomGPT''-style tools for network configuration, troubleshooting, and intent translation \cite{bariah2024next,zou2025telecomgpt,zou2026nree}. While these developments have significantly advanced the potential of such AI tools in telecom networks, they still experience key limitations pertinent to the fact that they are fundamentally pattern learners operating over sequences of tokens and tool calls, and hence, they lack the knowledge on the representation of the physical and logical dynamics of a communications network. This means that LLMs and LLM-based agents can interpret logs, generate configuration templates, and perform calls to external simulators and tools, but they are unable of capturing how network evolves, including the evolution of radio conditions, queues, mobility, and protocol states as a response to control decisions.

In parallel, the last decade has witnessed accelerated developments in Digital Twins (DTs) as virtual replicas of physical networks, comprising measurements, simulations, and visualization tools to support planning, testing, and what-if analysis over radio, transport, and core. DTs offer scenario-based view of the network, while relying on predefined propagation and traffic models, rendering them effective tools for emulating a wide range of scenarios and validating configurations. However, most current DTs are not learned, action-conditioned world models, in the sense that they do not learn system dynamics directly from the data, and they rely on fixed assumptions instead of explicitly modeling uncertainty. In addition, agents are rarely incorporated in DTs, limiting their capabilities to optimize decisions under changing network conditions.

On the other hand, wireless networks are characterized by their highly-dynamic multi-layered nature, that intertwines propagation and interference, user mobility patterns, traffic variations, protocols interactions, and policies. In specific, decisions taken at one layer, such as RAN scheduling, slicing policies, or routing, propagate through the multiple layers in the system and may translate into quality-of-service, energy, or reliability requirements. Accordingly, traditional systems in which automation is triggered by a key performance indicators (KPIs) degradation or a raised alarm, is becoming increasingly inadequate in dense, heterogeneous, and mission-critical 6G scenarios. This calls for models that are not purely reactive, in order to enable operators to anticipate congestion, failures, and risk, and to evaluate the impact of control actions before they are applied to the live network.

In the broader ML and robotics communities, world models (WMs), which are learned representations of environment dynamics that simulate system evolution under candidate actions, have emerged as a central abstraction for modeling and predicting system behavior~\cite{hafner2023dreamerv3, sora2024, lecun2022jepa}. Several paradigms exist (detailed in Section~\ref{sec:bg_wm}), but none natively supports the structured, multi-layered dynamics of telecommunication systems. Some initial attempts to integrate world models in the telecom networks have been developed. In specific, latent-dynamics world models have been explored for wireless resource control and edge intelligence, where internal models of the radio environment and traffic dynamics are used to improve reinforcement learning and planning~\cite{feriani2021wm_wireless, park2024mbrl_channel_access}. Furthermore, generative, field-oriented models have been proposed for mobile networks, learning to generate coverage, interference, and traffic fields conditioned on infrastructure configurations and environmental context~\cite{mobiworld2025}. At the same time, telecom-specific foundation models and large wireless models have been introduced as adaptable, task-agnostic representations of channels, traffic, and network behavior \cite{zhu2025wireless, shahid2025large}. These works provide valuable building blocks, but they remain largely specialized, one world model per use case, one DT per domain, and one foundation model per modality scenario.

This article argues that the future 6G systems require a more unified construct, which we refer to as the \textbf{Telecom World Models (TWMs)}. By TWM, we refer to a learned, action-conditioned model of how a telecom system evolves, jointly capturing the controllable aspects of the network, such as configuration, topology, and policies, and the external factors, including propagation, mobility, traffic, and failures. The model is designed to simulate possible future states, evaluate candidate actions, and support proactive decision-making. 
It is worth asking whether a DT can be considered a form of world model. At a high level, both aim to represent network behavior and support what-if analysis, and some advanced DT frameworks already incorporate DRL, surrogate models, or generative components. However, in this article, we use the notion of a world model in a more specific, planning-oriented sense, namely, a learned, action-conditioned predictive model that supports fast evaluation of alternative decisions under uncertainty. From this perspective, DTs and TWMs are not competing approaches but complementary ones. DTs provide high-fidelity environments for mirroring, validation, and data generation, while TWMs offer a learned predictive model that enables faster planning and decision-making within the control loop.

The main contributions of this article are as follows:
\begin{enumerate}
  \item A unified \textbf{\emph{two-world, three-layer}} architecture for telecom world models (TWMs), proposed as a factored-state Partially Observable Markov Decision Process (POMDP) with Controllable System World (CSW) / Exogenous World (EGW) decomposition. The architecture comprises a Field World Model~(FWM) for physical-layer dynamics, a Control/Dynamics World Model~(CDWM) for network-level control loops, and a telecom foundation model for intent translation and reasoning.
  \item Detailed designs for each layer, including representative tasks, key requirements, and inter-layer interaction patterns that ground the architecture in telecom-specific challenges spanning from electromagnetic field inference to intent-driven planning.
  \item A comparative analysis showing that TWM is the first construct to jointly provide telecom state grounding, fast action-conditioned rollouts, calibrated uncertainty, multi-timescale dynamics, model-based planning, and LLM-integrated guardrails.
  \item A practical methodology for building and evaluating TWMs, covering layer-specific data pipelines, training approaches, and a three-axis evaluation framework, including field accuracy, trajectory accuracy, and decision accuracy, paving the way for future benchmarks of TWMs.
  \item A proof-of-concept on multi-domain network slicing that validates the architecture, the full three-layer pipeline outperforms single-world baselines and budget-matched DT search across all tested scenarios.
  \item An outline of open challenges for TWMs spanning theory and foundations, algorithms and training, systems and deployment, safety, governance, and trust for production-grade TWM realization.
\end{enumerate}

In the remainder of the paper, we formalize the notion of a TWM, introduce a unified architecture based on two worlds and three layers, compare this construct with LLM agents, DTs, and existing world-model paradigms, discuss how a TWM can be built and evaluated in practice, and illustrate its potential through representative use cases and a proof-of-concept demonstration.

\section{Background and Related Work}
\label{sec:background}

In this section, we survey the key four pillars relevant to TWM concept, including world models in the broader ML community, Network Digital Twins (NDT), telecom-specific foundation models, and emerging telecom world-model-like efforts. For each, we summarize the core ideas and highlight the limitations that motivate a unified TWM architecture.

\subsection{World Models: Five Representative Paradigms}
\label{sec:bg_wm}

In the machine learning literature, a world model is a learned, generative model of environment dynamics that allows an agent to simulate, in its internal representation, how the world will evolve under candidate actions~\cite{ha2018worldmodels}. By forecasting future trajectories before acting, the agent can plan, explore safely, and learn with fewer interactions with the real environment. Within this context, five representative paradigms of world models have emerged over the past decade.

\textbf{Latent Dynamics Models}
The first and most mature models family focuses on learning a compact latent state together with an action-conditioned transition model. PlaNet~\cite{hafner2019planet} introduced a Recurrent State-Space Model (RSSM) that plans entirely in latent space. The Dreamer framework~\cite{hafner2020dreamer, hafner2023dreamerv3} extended this idea by learning a value function and a policy within the latent world model, achieving sample-efficient reinforcement learning across diverse control tasks. MuZero~\cite{schrittwieser2020muzero} proposed a complementary approach by coupling learned latent dynamics to online search, while more recent works have replaced the recurrent backbone with transformers~\cite{micheli2023iris} or combined learned dynamics with temporal-difference planning~\cite{hansen2022tdmpc}. A common factor among all these approaches is the use of latent trajectory simulations for counterfactual evaluation, together with uncertainty estimates induced by stochastic latent variables, ensembles, or planning-time search. These models are highly effective for single-agent, single-timescale control tasks with image-based observations. However, they do not inherently support multi-layered, multi-timescale, partially observable systems with structured state spaces, as in telecom networks.

\textbf{Generative and Video-prediction World Models}
A second family frames world modeling as predicting future observations in perceptual space. Early stochastic video generation methods~\cite{denton2018svg} learned to produce diverse plausible futures by combining variational autoencoders with recurrent networks. More recently, large-scale generative models have dramatically expanded the fidelity and duration of synthesized videos. Sora~\cite{sora2024} demonstrated that diffusion transformers trained on large-scale video data can generate long, coherent video sequences with consistent physical behavior. UniSim~\cite{yang2024unisim} and Genie~\cite{bruce2024genie} added action-conditioning, enabling interactive simulation of real-world and game environments respectively, while GameNGen~\cite{valevski2024gamengen} showed that diffusion models can serve as real-time game engines. These approaches produce rich and interpretable predictions and can capture complex visual dynamics, but they have three main limitations for telecom: i) the computational cost is extremely high, ii) the output modality (pixels or video frames) does not directly correspond to the structured KPIs and counters that operators need, and iii) action-conditioning is typically limited in granularity and is not designed for the combinatorial policy spaces of network management.

\textbf{Implicit World Models}
The third family learns predictive representations fully in the latent space, without reconstructing observations explicitly. The key idea is that a model can capture environment dynamics by predicting future latent embeddings rather than future pixels or signals, and hence, avoiding the cost of a decoder while retaining abstract and semantically meaningful structure. LeCun's Joint Embedding Predictive Architecture (JEPA)~\cite{lecun2022jepa} formalizes this approach, and V-JEPA~\cite{bardes2024vjepa} adapts it for video by predicting masked spatio-temporal regions in embedding space. These models produce reusable representations and are computationally lighter compared to their generative counterparts. However, they lack explicit action-conditioning, do not produce calibrated probabilistic forecasts over system outcomes, and cannot perform the multi-step rollouts required for planning under uncertainty.

\textbf{VLA and Multi-modal Embodied World Models}
The fourth emerging family integrates visual perception, language-conditioned intent, action generation, and predictive world modeling in embodied environments. 3D-VLA~\cite{zhen2024threedvla} introduced a generative world model that links 3D perception, reasoning, and action through language-conditioned embodied interaction, while WorldVLA~\cite{cen2025worldvla} unified vision-language-action modeling with autoregressive world prediction in a single framework. These models are important because they depart from the classical ``state-to-action'' robotics pipeline, by jointly modeling instructions, perception, action, and predicted future observations. In telecom, this approach is conceptually significant because it is the closest to coupling a foundation model reasoning layer with an action-conditioned predictive model. Its limitation is that the underlying environment remains focused on embodied manipulation or navigation rather than KPI-centric, constraint-heavy network operations.

\textbf{3D World Models}
The fifth family explicitly models spatially grounded, geometry-aware world evolution, instead of 2D pixels or compact latent tokens. OccWorld~\cite{zheng2024occworld} learned a 3D occupancy world model for autonomous driving that predicts both scene evolution and ego motion, while GeoDrive~\cite{chen2025geodrive} incorporated explicit 3D geometry into a driving world model for more precise controllable future predictions. Together with works such as 3D-VLA~\cite{zhen2024threedvla}, these models show that modern world modeling is increasingly concerned with structured physical space, not only perceptual appearance. Hence, this is more oriented to telecom than game-playing or image-centric settings because wireless behavior also depends on geometry, coverage, blockage, topology, and spatial demand. However, existing 3D world models still target robotics or driving scenes rather than action-conditioned forecasting of telecom KPIs under operator policies.

\textbf{Implications for Telecom}
Taken together, these five paradigms contribute with the following complementary ideas, 1) latent dynamics models provide compact action-conditioned rollouts, 2) generative perceptual models provide rich future simulation, 3) implicit models provide abstract predictive representations, 4) VLA models provide instruction-conditioned action reasoning, and 5) 3D world models provide geometry-aware state evolution. While these feature provide advanced and different approaches for understanding the world, telecom systems combine characteristics that none of these can address individually, including structured KPI-centric state, multi-tenant and multi-agent coupling (for example across slices, services, and operators sharing the same infrastructure), control loops spanning milliseconds to hours, and operator intent expressed as policies, constraints, and service level agreements (SLAs), instead of just physical manipulation goals. Therefore, directly applying existing world-model architectures requires substantial adaptation in state representation, action spaces, uncertainty modeling, and training methodology. This gap motivates the design of telecom-specific world-model layers, as detailed in Sections~\ref{sec:fwm} and~\ref{sec:cdwm}. While world models address the learning and planning side of the problem, the telecom community has progressed in a parallel path through simulation-based digital replicas and domain-specific foundation models. In the following, we examine these approaches.

\subsection{Network Digital Twins}
\label{sec:bg_dt}

A Network Digital Twin (NDT) is a virtual replica of a network physical infrastructure that reflects in a synchronous way its current state, supports what-if analysis, and can be used for planning, testing, and optimization~\cite{itut2020y3090}. NDTs have received growing attention from the telecom community, including standardization bodies (ITU-T Y.3090~\cite{itut2020y3090}, O-RAN), industry, and academia~\cite{khan2022dt, almasan2022dt, khan2022dt_wireless}, as enablers for intelligent 5G-Advanced and 6G management. A typical NDT comprises three main elements, data collected from a live network, a simulation engine (ray-tracing, system-level, and transport models), and a visualization and analytics layer.

Despite their potential, current NDTs exhibit four limitations relevant to the vision of TWM, namely, they are primarily descriptive instead of learned, scenario-based (requiring same simulation runs for all candidate policies), treat uncertainty implicitly through Monte Carlo repetition, and are rarely coupled with intent-aware control systems. These gaps motivate the CDWM layer of the TWM as a learned model that enables fast, action-conditioned, and uncertainty-aware predictions while maintaining DT-level fidelity.

\subsection{Telecom Foundation Models and TelecomGPT}
\label{sec:bg_fm}

The success of LLMs has led to the development of telecom-specific foundation models~\cite{hao2025CST,zou2026nree}, including TelecomGPT~\cite{zou2025telecomgpt}, NetGPT~\cite{netgpt2024}, WirelessLLM~\cite{wirelessllm2024}, RF-GPT~\cite{zou2026rf} and wider explorations of generative AI for telecom~\cite{bariah2024next, boateng2024llmNSM, long2024llm6g, liang2025llmWireless}. In parallel, large wireless models learn reusable representations from signals, channels, and KPI time series~\cite{zhu2025wireless, shahid2025large}. Key use cases cover configuration generation, troubleshooting, intent translation, and question answering over standards~\cite{maatouk2024teleqna}, with recent agentic approaches orchestrating tools and simulators through multi-step reasoning~\cite{chatnet2023}.

Despite these advances, telecom foundation models remain reactive pattern matchers, in the sense they do not maintain an internal, action-conditioned model of network dynamics, and they cannot produce calibrated KPI distributions. Furthermore, they are highly susceptible to hallucination in safety-critical settings~\cite{hao2025CST}. It is important to note that these models perform well in language-level reasoning, tool orchestration, and knowledge retrieval. However, they do not capture predictive system dynamics, which in TWM is handled by the FWM and CDWM layers. A number of recent efforts have begun to introduce learned predictive components into telecom systems, even if they are not framed as world models.

\subsection{Existing Telecom World-Model-like Efforts}
\label{sec:bg_telco_wm}

Although the term ``world model'' is rarely used in the telecom literature, several recent frameworks can be interpreted as partial realizations of this concept. In the following, we categorize them into groups.


\textbf{Latent Models for Radio Resource Management (RRM) and Network Control.}
Model-based RL has been explored for wireless resource allocation~\cite{eisen2019mbrl}, sample-efficient wireless control~\cite{feriani2021wm_wireless}, channel access~\cite{park2024mbrl_channel_access}, latent multimodal dynamics for optimized resource planning~\cite{chaaya2026twc}, and foundation-model-scale latent dynamics of mobile network state~\cite{mobiworld2025}. These efforts demonstrate the value of learned dynamics but remain largely scoped to individual control loops, limited planning horizons, or specific modalities, without jointly integrating spatial field awareness, multi-timescale control, and intent-aware planning.

\textbf{Neural Field Generators.}
RadioUNet~\cite{levie2021radiounet}, PMNet~\cite{lee2023pmnet}, deep completion autoencoders~\cite{teganya2022radiomap}, and differentiable ray tracing~\cite{orekondy2023winert} learn spatial RF fields for coverage and interference estimation. They partially represent the FWM layer, but are typically single-task, do not provide action-conditioned control, and are not coupled to a dynamics or intent layer.

\textbf{Learned DT Surrogates.}
RouteNet~\cite{ferriol2022routenet} and similar graph neural network surrogates predict per-flow KPIs from network configurations, providing fast alternatives to packet-level simulation. However, they are single-step predictors without temporal understanding, explicit uncertainty, or coupling to intent-aware planning.

\textbf{Wireless Simulation and DT Platforms.} 
High-fidelity wireless simulation and DT platforms, such as Sionna/Sionna RT and NVIDIA Aerial Omniverse DT ~\cite{hoydis2022sionna,hoydis2023sionna,aoudia2025sionna,nvidia_aerial_dt_2026}, enable differentiable, site-specific, and large-scale network simulation, supporting what-if analysis, channel generation, radio-map construction, and synthetic data generation. Furthermore, AI-enhanced DT frameworks, such as DeepVerse \cite{demirhan2022deepverse}, integrate learning-based optimization into these environments. However, these systems are simulation-centric, which require explicit runs for each scenario, lack learned action-conditioned dynamics, and are not integrated with intent-aware planning.


\textbf{Safe and Constrained RL For Network Slicing.} 
Constrained reinforcement learning has been applied to network slicing with a focus on SLA safety during training and deployment. ~\cite{nagib2025safeslice} proposes a safe DRL framework for O-RAN inter-slice resource allocation, while ~\cite{zangooei2024flexible} studies constrained multi-agent RL for coordinated PRB allocation under capacity constraints. These approaches reduce SLA violations compared to unconstrained baselines. However, they remain model-free and cannot perform multi-step counterfactual predictions required for planning. Moreover, they rely on aggregate KPI observations without spatial conditioning, limiting their ability to anticipate localized effects, such as hotspot-driven congestion.

\textbf{Closed-Loop O-RAN Control Implementations.}  

Recent works have demonstrated end-to-end closed-loop slicing control on open-source O-RAN platforms. For example, the authors in ~\cite{barker2025real} present a framework combining the O-RAN Software Community near-RT RIC with srsRAN, where an RL-based xApp dynamically allocates PRBs across eMBB, URLLC, and mMTC slices via the E2 interface, with policy updates driven by real-time throughput and QoS feedback. Also, ~\cite{cheng2024oranslice} extends OpenAirInterface with a two-tier radio resource scheduler and an E2SM-CCC-based service model for O-RAN-compliant RAN slicing, validated on both software-defined radio testbeds and private 5G platforms. ~\cite{bonati2023openrangym} introduces OpenRAN Gym, a toolbox for data-driven O-RAN experimentation providing xApp skeletons, data pipelines, and integration with large-scale wireless emulators. However, these systems focus on execution instead of planning, where decision logic is rule-based or model-free, without learned dynamics, spatial awareness, or intent-level reasoning.

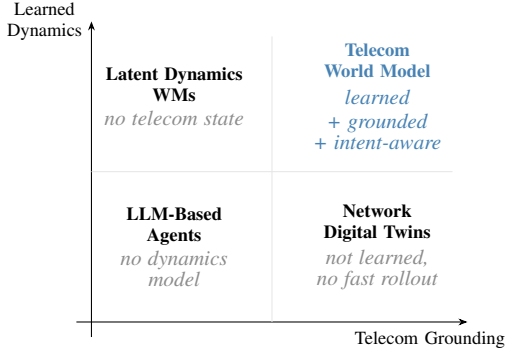
\begin{figure}[t]
\centering
\begin{tikzpicture}[font=\scriptsize]
  \draw[-{Stealth[length=3.5pt]}, thin] (-0.2,0) -- (5.0,0)
    node at (4.5, 0) [below, font=\scriptsize] {Telecom Grounding};
  \draw[-{Stealth[length=3.5pt]}, thin] (0,-0.2) -- (0,4.0)
    node[left, font=\scriptsize, align=center] {Learned \\Dynamics};
  \node[align=center, text width=20mm] at (1.1,3.0)
    {\textbf{Latent Dynamics\\WMs}\\[1pt]
     {\footnotesize\itshape\color{black!45} no telecom state}};
  \node[align=center, text width=22mm] at (3.8,1.0)
    {\textbf{Network\\Digital Twins}\\[1pt]
     {\footnotesize\itshape\color{black!45} not learned,\\no fast rollout}};
  \node[align=center, text width=20mm] at (1.1,1.0)
    {\textbf{LLM-Based\\Agents}\\[1pt]
     {\footnotesize\itshape\color{black!45} no dynamics model}};
  \node[align=center, text width=22mm, text=steelblue] at (3.8,3.0)
    {\textbf{Telecom\\World Model}\\[1pt]
     {\footnotesize\itshape learned \\+ grounded\\+ intent-aware}};
  \draw[black!10, thin] (2.4,0) -- (2.4,3.8);
  \draw[black!10, thin] (0,2.0) -- (4.8,2.0);
\end{tikzpicture}
\caption{Positioning of TWM in the landscape of related paradigms. Each approach occupies a different region of the telecom-grounding vs.\ learned-dynamics space. The TWM synthesizes learned dynamics, telecom-specific state, and foundation-model reasoning.}
\label{fig:landscape}
\end{figure}

\textbf{Gap Analysis.}
Given the aforementioned frameworks, we can observe a unified pattern across all approaches. In specific, world models target image-based, single-agent settings, NDTs are descriptive and costly, telecom foundation models lack internal dynamics, and existing telecom efforts address only parts of the problem (one model per control loop and one field generator per scenario).  None of these works provides a unified, multi-layer, action-conditioned, uncertainty-aware architecture covering electromagnetic fields to operator intent.  Fig.~\ref{fig:landscape} positions the TWM with relative to these paradigms, the architecture introduced in Section~\ref{sec:twm} is designed to fill this gap.

\section{Telecom World Model}
\label{sec:twm}

\subsection{Telecom World Model: Definition}

To transcend the intuitive notion of ``a brain inside the network’’, we define a TWM as a learned, action-conditioned model of telecom system dynamics, formalized as a POMDP with factored state. This definition builds on the broader world-model view of learned internal dynamics for imagination, planning, and counterfactual evaluation~\cite{ha2018worldmodels,hafner2019planet,hafner2023dreamerv3}. At time $t$, let $s_t \in \mathcal{S}$ denote the (hidden) telecom state, $o_t \in \mathcal{O}$ the observation available to the operator, and $a_t \in \mathcal{A}$ the control action. Also, $s_t$ comprises two components, corresponding to the controllable system world and the exogenous world, and is factored as $s_t=(s_t^{\mathrm{CSW}}, s_t^{\mathrm{EGW}})$. Hence, the TWM is characterized by the tuple $(\mathcal{S},\,\mathcal{O},\,\,\mathcal{A},\,T,\,R)$:

\begin{itemize}
  \item \textbf{State} $\mathcal{S} = \mathcal{S}^{\mathrm{CSW}} \times \mathcal{S}^{\mathrm{EGW}}$: the (hidden) telecom state, including infrastructure configuration, radio and transport resources, traffic information and queues, user mobility, channel and interference fields, and fault status.

  \item \textbf{Observations} $\mathcal{O}$: the measurements and traces available to the operator, including performance counters, Minimization of Drive Tests (MDT) reports, KPI and RF measurements, and telemetry from simulators or DTs. These are partial, noisy, and delayed projections of the hidden state.

  \item \textbf{Actions} $\mathcal{A}$: operator and control actions that influence the system, such as RAN configuration (power, scheduling, handover parameters), slicing and admission-control policies, routing decisions, and energy management actions.

  \item \textbf{Dynamics} $T(\mathbf{s}_{t+1} \mid \mathbf{s}_t,\, \mathbf{a}_t)$: the evolution of the state over time under actions and external influences, spanning timescales from millisecond-scale fading to sub-second scheduling and hour-scale traffic shifts and capacity planning. This joint transition decomposes into controllable and external components.

  \item \textbf{Objectives} $R(\mathbf{s}_t,\, \mathbf{a}_t)$: multi-objective utility combining Quality of Experience~(QoE), Quality of Service~(QoS), resource and energy efficiency, operational cost, and SLA compliance.
\end{itemize}


The key structural characteristic of the TWM is its \emph{asymmetric factored dynamics}, in which the telecom state is decomposed into a controllable system world and an exogenous environment, and operator actions directly affect only the former. The exogenous world evolves as a stochastic process conditioned on the joint state rather than being directly actuated by the operator. Following factored MDP/POMDP representations based on dynamic Bayesian networks~\cite{hansen2000factored_pomdp}, we adopt the following telecom-specific factorization of the transition model,
\begin{equation} \small
T(\mathbf{s}_{t+1} \mid \mathbf{s}_t,\, \mathbf{a}_t)
\;=\;
T^{\mathrm{CSW}}\!\bigl(\mathbf{s}_{t+1}^{\mathrm{CSW}} \mid \mathbf{s}_t,\, \mathbf{a}_t\bigr)
\;\cdot\;
T^{\mathrm{EGW}}\!\bigl(\mathbf{s}_{t+1}^{\mathrm{EGW}} \mid \mathbf{s}_t\bigr).
\label{eq:factored_dynamics}
\end{equation}

Eq.~\eqref{eq:factored_dynamics} is represented as a modeling abstraction, not a strict separation. It is reasonable when operator actions primarily affect the network state and external factors evolve independently, but the boundary can fade over longer timescales due to feedback effects. Accordingly, this factorization captures three properties that distinguish the TWM from a generic POMDP, namely, (i)~the action dependency is asymmetric, meaning that the operator controls the CSW given environmental context, while the EGW (propagation, mobility, traffic, failures) evolves independently of the action, (ii)~the two worlds couple through state, not through actions, enabling modular modeling, and (iii)~the dynamics are action-conditioned on the CSW side, allowing the model to evaluate alternative actions/policies and predict their impact on future system states.

Beside these core elements, uncertainty is one hidden element of TWM and can be observed in three forms. First, inherent randomness in propagation, traffic, and user behavior means that even a perfect model cannot predict the future in a deterministic way. Second, partial observability means that the operator can never see the full network states, but only incomplete, noisy, and delayed measurements, so the TWM must reason over anticipations rather than known states. Third, model uncertainty arises because any learned model is trained on finite data and experiences approximation errors, so the TWM must quantify its own confidence and propagate it through the planning pipeline. These three forms of uncertainty enable the TWM to be planning-ready and not only predictive model. 

\begin{figure}[t!]
\centering
\includegraphics[width=1.0\linewidth]{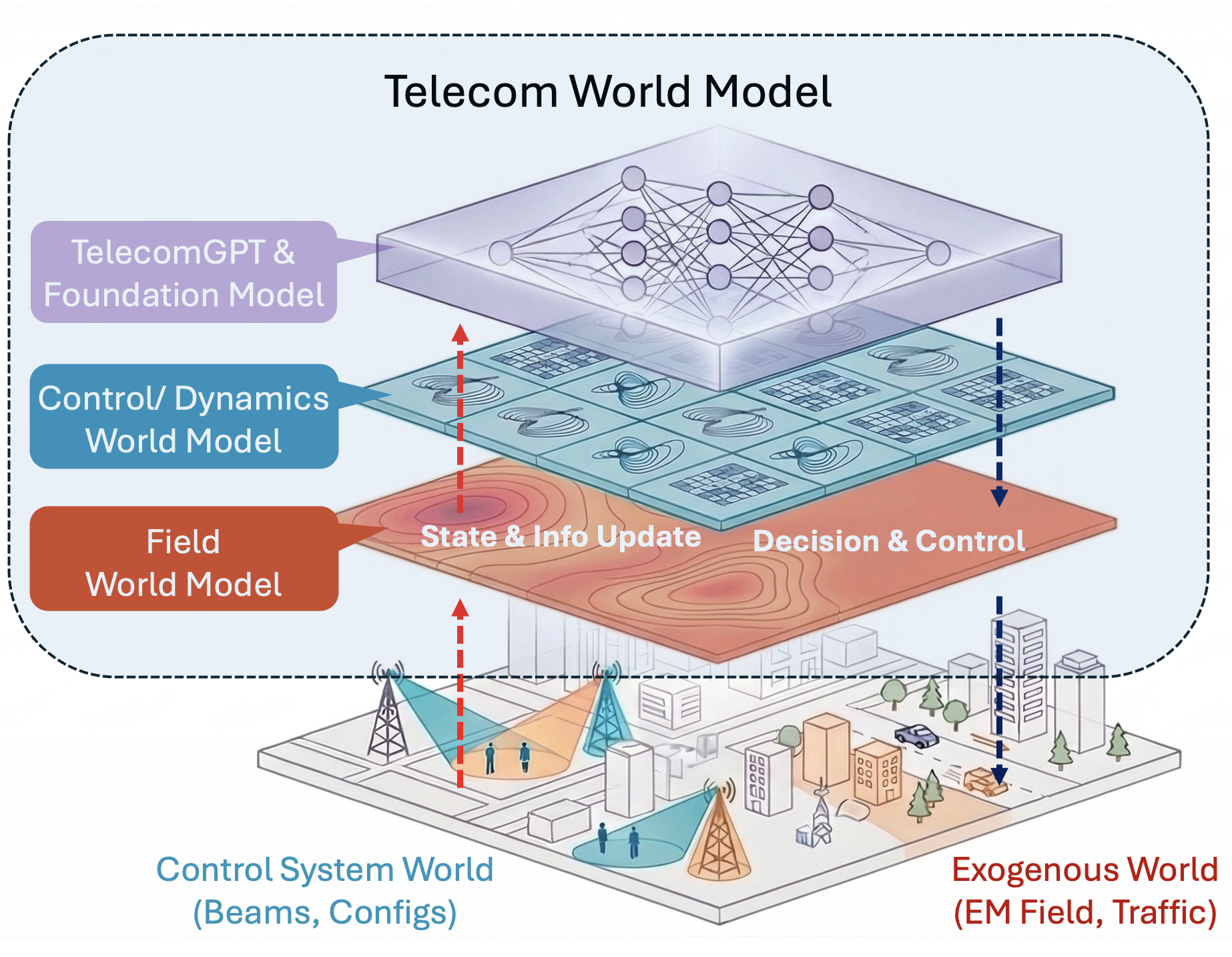}
\caption{Two-world, three-layer Telecom World Model (TWM), showing CSW/EGW factorization and FWM--CDWM--TelecomGPT interaction loops.}
\label{fig:TWM}
\end{figure}

The factored state $\mathcal{S} = \mathcal{S}^{\mathrm{CSW}} \times \mathcal{S}^{\mathrm{EGW}}$ and the asymmetric transition in Eq.~\eqref{eq:factored_dynamics} are the key backbone of the TWM. In the following subsection, we propose the detailed definition of each world, before we move for the full description of the three architectural layers that complement them. Fig.~\ref{fig:TWM} provides an overview of the two-world, three-layer architecture that shapes the TWM.

\subsection{Two Worlds: Controllable System and Exogenous Environment}

A central feature of telecommunications, compared to other benchmark environments in the open literature, is the clear separation between what the operator can control directly and what is impacted by external, stochastic factors. For the purpose of proposing a clear definition of a TWM, we introduce two distinguished conceptual ``worlds’’ that interact with each others, but each plays a fundamentally different role.

The first is the \textbf{\emph{Controllable System World (CSW)}}. This comprises all aspects of the network that can be configured or modified by the operator or by the corresponding automation systems. This world comprises the logical and physical topology of the infrastructure, radio and transport configurations, network slicing and admission policies, scheduler and handover parameters, routing and load-balancing strategies, and higher-layer policies that control energy management and service orchestration, among other parameters. These elements are represented in the control and dynamics part of the TWM and constitute the mechanism through which the operator can influence the system behavior.

The second is the \textbf{\emph{Exogenous World (EGW)}}. This captures phenomena that are not directly controllable, but fundamentally affect the network performance and must be modeled to reason about future behavior. It encompasses radio propagation and shadowing due to terrain and buildings, user and device mobility patterns, spatial and temporal traffic demand, external interference, and spontaneous failures or degradations in the environment. These processes are inherently stochastic and only partially observable through measurements and reports. It also includes action-bearing external agents whose behavior is not controlled by the operator, for example, neighboring cells or networks, user/application adaptation, and other third-party systems that introduce additional demand, interference, or failure dynamics. In the TWM architecture, these parameters are captured by the field world model (FWM) (detailed in the following subsection), which learns spatial and spatio-temporal fields, including coverage, interference, load, and mobility maps, conditioned on the system state and context. 

The interaction between the CSW and the EGW is at the heart of the TWM, in which the control actions act on the former, their consequences propagate through the both, and the latter states impact the former actions. This separation is also consistent with digital-twin thinking, where operator-configurable network state is distinguished from traffic, propagation, and environmental context~\cite{itut2020y3090,khan2022dt,almasan2022dt}. Within this cycle, the TWM must capture this coupled evolution in order to support reliable prediction, planning, and optimization.

The CSW and EGW decomposition is motivated by three design principles. First, \textbf{controllability}, the CSW–EGW split distinguishes between what the operator can control and what it must adapt to, which is essential for formulating meaningful optimization objectives and constraints. Second, \textbf{modular learnability}, where the exogenous world, driven by physics, geography, and user behavior, can be learned from spatial and environmental data, while the controllable system, shaped by protocols and policies, is better modeled using action-conditioned sequence models. Third, \textbf{interpretability}, where operators need to understand why a recommendation is made, as separating environmental conditions from policy choices makes it easier to attribute predicted outcomes.

\subsection{The Three Layers: Field WM, Control/Dynamics WM, and TelecomGPT}

Building on the definition and roles of the CSW and the EGW, we structure the TWM into three interacting layers. Each layer targets a distinct aspect of the problem and operates at an appropriate level of abstraction, while exposing clear interfaces to the others.

The \textbf{\emph{Field World Model (FWM) layer}} focuses on the exogenous world. It learns spatial, spatio-temporal, and electromagnetic (EM) fields that characterize the environment in which the network operates, such as coverage and signal-to-interference patterns across a geographical area, traffic density and load maps, and user or vehicle mobility fields. Given a static context, like terrain, building layouts, and the current or candidate configuration of the network, this layer produces probabilistic forecasts of these fields over time, hence, capturing how the environment is likely to evolve under different conditions. By modeling these quantities explicitly, the FWM provides an interpretable, uncertainty-aware view of propagation, mobility, and traffic that can be used both for planning and for validating decisions. Technically, this layer is closest to learned radio-map, pathloss-map, and differentiable wireless propagation models~\cite{levie2021radiounet,lee2023pmnet,teganya2022radiomap,orekondy2023winert}.

The \textbf{\emph{Control/Dynamics World Model (CDWM) layer}} focuses on the controllable system world. It represents the evolution of the internal network state and KPIs under the operator actions and external influences. This layer typically operates in a latent space, where it learns compact representations of configuration, load, queues, slice states, and other internal variables, together with their action-conditioned dynamics across relevant timescales. Its role is to predict the KPIs evolution under different control policies, enabling fast evaluation of candidate actions without running full simulations. In combination with the FWM, it links high-level control decisions to both overall performance metrics and spatially resolved effects. Conceptually, the CDWM sits at the intersection of latent-dynamics world models and learned network-performance surrogates adapted to telecom state and action spaces~\cite{hafner2023dreamerv3,micheli2023iris,ferriol2022routenet,mobiworld2025}.

The third layer is the \textbf{\emph{TelecomGPT/Foundation Model layer}}. This layer is realized by a telecom-specific foundation model or an LLM that serves as the cognitive interface to the TWM and to the broader network system. It acts as an intent translator, in order to transform high-level requirements and policies expressed by operators, in natural language or in structured intent formats, into machine-readable objectives and constraints. It also plays the role of tool orchestrator, where it coordinates calls to the FWM and CDWM, external simulators and DTs, O-RAN RIC xApps and rApps, policy engines and constraint solvers, and observability and analytics services. On top of this, the TelecomGPT layer performs planning and explanation, in which it compares alternative futures generated by the underlying world models, applies SLA and safety constraints, recommends actions and policies, and explains its decisions in a human-readable way. In this way, the foundation-model layer turns the predictive capabilities of the TWM into actionable, operator-aligned intelligence. This layer builds directly on recent telecom-specialized foundation models and agentic tool-orchestration frameworks~\cite{zou2025telecomgpt,netgpt2024,wirelessllm2024,chatnet2023}. \textit{It is worthy to note that, in our current work, the TelecomGPT/foundation-model layer is not treated as a standalone language model, but as an agentic orchestration layer, responsible for translating intents, querying the underlying world models, coordinating tool calls, and enforcing constraints to enable model-grounded planning.}

\begin{figure}[t]
\centering
\includegraphics[width=\columnwidth]{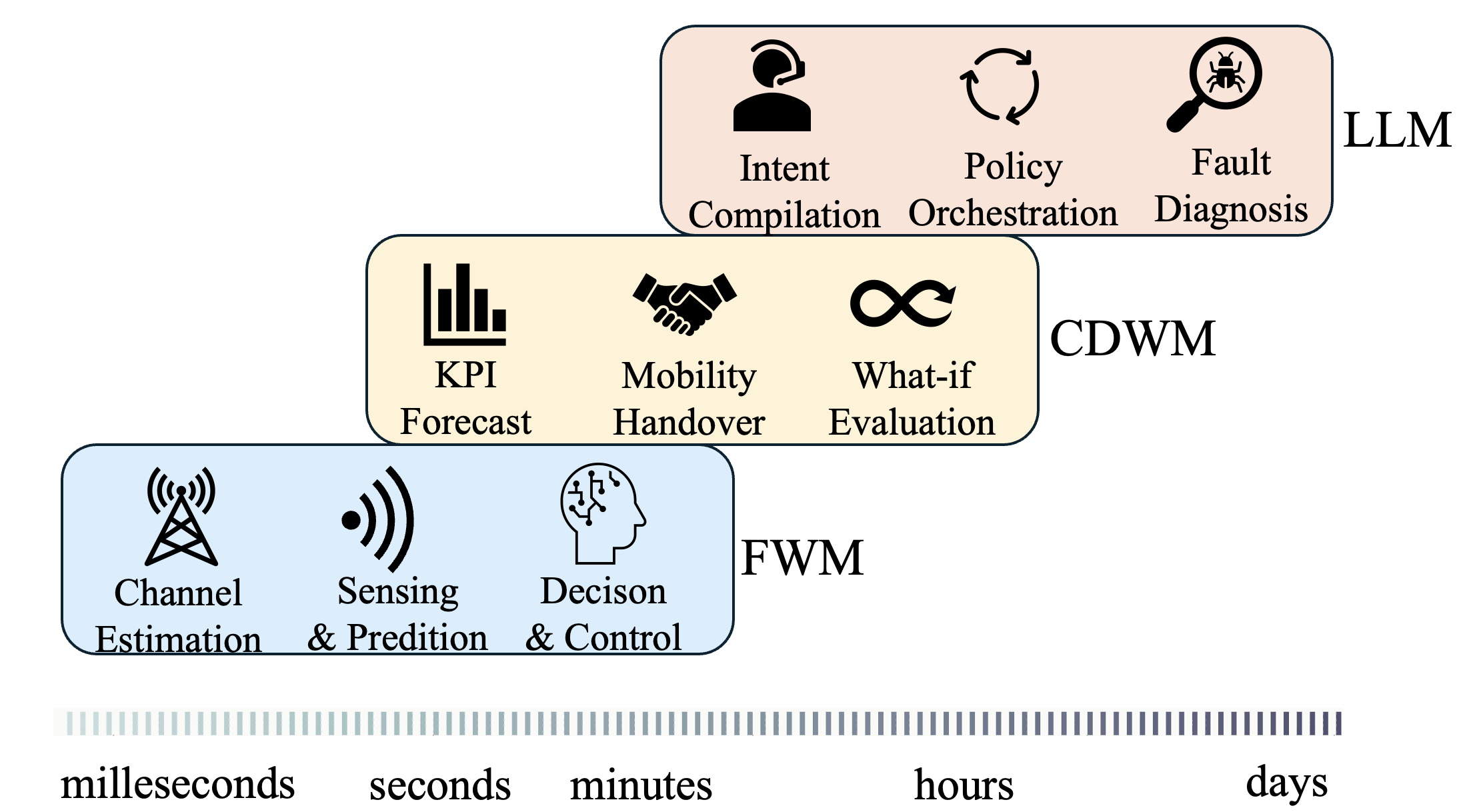}
\caption{TWM layers across planning timescales.  Each horizontal band represents the temporal regime in which one layer dominates, with representative tasks.  The overlapping bands reflect the paper's argument that the layers share timescale boundaries rather than partitioning them cleanly.}
\label{fig:timescales}
\end{figure}

The separation between the FWM and the CDWM should not be interpreted as a strict PHY versus network split, but instead, it reflects the differences in control granularity, timescale, and scope. Within this context, we define \textbf{micro-actions} as fast, local decisions taken at the radio-facing interface of the network, such as beam updates, Modulation and Coding Scheme (MCS) selection, instantaneous scheduling adjustments, and handover trigger execution. We also define \textbf{macro-actions} as slower, policy-level decisions that configure the behavior of local controllers over a planning horizon, such as slice resource budgets, scheduler weights, mobility parameter sets, admission thresholds, cell on/off decisions, and power or tilt envelopes. In this sense, the primary data flow goes from FWM to CDWM, as follows. The FWM produces spatial context (coverage, load, interference features) that the CDWM consumes as conditioning inputs for its KPI trajectory predictions. Thus, the FWM forms the inner loop of the TWM, capturing short-horizon interactions with the field environment, while the CDWM forms the outer loop, predicting system-level KPI trajectories under candidate macro-actions. Feedback from the CDWM to the FWM occurs when candidate configurations modify the spatial operating conditions of the system. Accordingly, the same telecom function might appear in both layers at different abstraction levels. For example, handover trigger execution belongs to the micro level, while handover policy tuning belongs to the macro level. Fig.~\ref{fig:timescales} illustrates how the three components operate across different, but overlapping, planning timescales, from millisecond-scale radio response to seconds-to-hours control dynamics and operator-level strategic planning.

\subsection{The Interactions between the two worlds and the three layers}

The TWM is not a collection of separate components, but its high potential comes from the way the FWM, the CDWM, and the TelecomGPT layer interact to connect the CSW and the EGW. In this subsection, we outline the main interaction patterns.

First, the FWM and the CDWM exchange information in both directions. The FWM produces spatio-temporal/EM fields describing coverage, interference, load, and mobility. These fields can be summarized into spatial statistics and contextual features. For example, distributions of signal quality, identification of hot-spots and coverage holes, or mobility patterns across different cells, which are then provided as and input to the CDWM. This allows the dynamics layer to condition its predictions of KPIs evolution on the structure of the underlying environment, rather than operating only on aggregate metrics. On the other hand, the CDWM generates trajectories of configurations and policies, such as cell on/off decisions, scheduling and handover parameters, or slice resource allocations, under candidate control strategies. These trajectories are fed back to the FWM to condition its forecasts of future fields, ensuring that environmental predictions remain consistent with the evolving system state.

Second, both world-model layers interface with the foundation-model layer. For a given intent or control problem, TelecomGPT queries the CDWM and the FWM to generate \emph{candidate futures}, including predicted KPI trajectories and spatial field evolutions under different action sequences. The outputs of the two world models, such as quantitative KPI predictions, uncertainty estimates, and spatial maps, are formatted into a representation that the foundation model can reason over. In particular, the world models provide TelecomGPT with a planning foundation that is grounded in telecom physics and network dynamics, instead of just textual context.

Third, TelecomGPT closes the loop by translating operator intents into objectives and constraints, and by orchestrating tools and world models accordingly. It interprets high-level requirements that are expressed in natural language, maps them to optimization criteria and feasible action sets, and leverages these tools to evaluate and select candidate actions. As part of this orchestration, TelecomGPT ensures that proposed actions comply with SLA, regulatory, and safety constraints, with rollback or safe fallback mechanisms applied when needed. Finally, it returns recommended actions or policies to the operator or automation system, together with explanations that link these recommendations to the underlying world-model predictions and constraints.

Through these interactions, the CSW and the EGW are coupled, and the three components of the TWM jointly enable prediction, planning, and safe decision-making in telecom networks. The end-to-end information flow and layer responsibilities are summarized in Fig.~\ref{fig:TWM}.

\section{Field World Model}
\label{sec:fwm}

As discussed earlier, the FWM focuses on short-horizon interactions between the radio system and the physical environment, enabling the modeling and forecasting of the spatial, spatio-temporal, and electromagnetic fields, such as channel state, coverage, interference, blockage, and mobility-sensitive radio conditions. In the hierarchical TWM architecture, it supports micro-control, in order to enable fast local decisions that react to instantaneous observations under the macro level defined by the CDWM. Hence, the FWM does not perform network-wide planning, however, it provides the inner-loop predictive and reactive capabilities for fine-grained radio–environment interactions. This role is anticipated by recent work on radio-map estimation, completion models, pathloss prediction, and differentiable neural ray tracing~\cite{levie2021radiounet,teganya2022radiomap,lee2023pmnet,orekondy2023winert,yan2025reading}.

As the basis of the large telecom system, the FWM focuses on the local tasks specified by the CDWM and provides feedback based on local observations. Within this context, it handles detailed radio-level interactions, in order to allow higher level models to operate without directly processing low-level electromagnetic signals. Therefore, it represents the fundamental component of the overall system, directly determining the performance of all upper-level tasks. Moreover, the FWM processes raw observations at high speed, hence, enabling it to sense and report changes in the physical environment in real time basis.
\subsection{Representative Tasks of the FWM}

\begin{table*}[t]
    \begin{tabular}{lll}
        \toprule
        \textbf{Category} & \textbf{Task} & \textbf{Description} \\
        \midrule
        \multirow{4}{*}{Channel Estimation} &
        Compression \& feedback &
        Compress high-dimensional CSI and enable efficient feedback to the transmitter. \\
        &
        Denoising \& interpolation &
        Recover channel through noisy and incomplete observations. \\
         &
        Prediction \& mapping &
        Forecast short-term channel evolution across time and frequency. \\
        &
        Overall estimation &
        Generally consider all available predictions and observations to generate high-quality CSIs.\\
        \midrule
        \multirow{3}{*}{Sensing \& Prediction} &
        User and object localization &
        Infer positions or trajectories of users and surrounding objects from RF signals. \\
         &
        Environment sensing &
        Build a dynamic model of targets in the physical environment from RF signals. \\
         &
        Mobility \& blockage prediction &
        Predict future mobility patterns and potential blockage events based on sensing. \\
        \midrule
        \multirow{5}{*}{Decision \& Control} &
        Initial access \& cell search &
        Accelerate access procedures by exploiting learned world representations. \\
         &
        Beamforming \& beam tracking &
        Select and update beams based on reconstructed and predicted channel states. \\
         &
        Interference management &
        Allocate resources and coordinates links to mitigate inter-cell and intra-cell interference. \\
         &
        MCS \& link adaptation &
        Adapt modulation and coding schemes according to link quality and reliability indicators. \\
         &
        Handover \& mobility control &
        Trigger and optimize handover decisions for moving users through mobility prediction. \\
        \bottomrule
    \end{tabular}
    \centering
    \caption{Representative Tasks of the FWM}
    \label{tab:fieldlayer}
\end{table*}
In this subsection, we present the defining tasks of the FWM. Generally, the field tasks can be divided into three categories, including channel estimation, physical resource management, and scenario prediction.

\textbf{Channel estimation, interpolation, and mapping:}
Channel-related tasks are essential for physical AI. In MIMO-OFDM systems, the channel matrix is high-dimensional and represents the sparse physical environment, hence, its reconstruction from partial and noisy observations is very essential. AI-based methods are employed to denoise, interpolate, and predict channel representations by exploiting their underlying structure. Given that the modeling of the physical world in explicit way is infeasible, a world model is introduced to implicitly capture environmental dynamics, enabling channel reconstruction within target resource blocks and improving transceiver throughput.

\textbf{Sensing and prediction:}
Sensing and prediction rely on inferring latent environmental dynamics and their temporal evolution. Although such mappings are highly compressive and cannot fully recover the physical world from wireless signals, world models enable implicit perception through high-dimensional observations and the learning of correlations across time, frequency, and space. This internal representation captures the mobility, blockage, and channel variations, enabling proactive communication, reducing pilot overhead, and improving robustness.

\textbf{Beamforming, initial access, interference management, MCS selection, and handover:}
At the highest level of the FWM, decision-making tasks, such as beamforming, initial access, interference coordination, MCS selection, and handover, leverage reconstructed channels and predicted environmental states to determine system actions. Unlike traditional modular designs based on predefined manual rules, AI-enabled approaches jointly optimize these decisions by learning mappings from channel and sensing outputs to system-level actions. Integrated within the FWM, these tasks require immediate responsiveness to environmental changes, enabling adaptive operation across diverse scenarios and improving spectral efficiency, reliability, and throughput.

Aforementioned field tasks are summarized in Table \ref{tab:fieldlayer}.

\subsection{Key Requirements of the FWM Layer}
As the FWM layer needs to deal directly with frequent and intense changes in the physical world in real time, additional challenges arise in understanding the world and responding accordingly.

\textbf{Handling stochastic and partial observations:}
Wireless systems inherently operate with stochastic and partial observations, due to noise, limited pilots, quantization, and hardware impairments, where only incomplete projections of the physical world are available. Instead of relying only on statistical assumptions, world models learn structured latent representations that capture both deterministic and random components. This allows the FWM to infer missing information, quantify uncertainty, and improve channel estimation, sensing, and decision-making under imperfect observations.

\textbf{Low computational complexity and instant decision:}
FWM tasks face strict real-time constraints imposed by channel dynamics and user mobility, making computational efficiency very critical. Although complex models may offer high accuracy, they are generally impractical for deployment under stringent latency requirements. Therefore, world models must balance expressiveness and efficiency by compressing high-dimensional interactions into compact latent representations, enabling fast inference and low-latency decision-making, without relying on excessive model scaling.

\textbf{Feedback for upper layers:}
The FWM layer also provides essential feedback to upper layers by summarizing its understanding of the physical environment into structured representations, such as reliability metrics, environmental dynamics, and predicted link quality. This enables cross-layer coordination and long-term optimization, bridge the gap between fast physical-layer reactions and higher-level strategies, and transforms the communication system into a unified, adaptive hierarchy built upon a shared world representation.
\section{Control/Dynamics World Model}
\label{sec:cdwm}

While the FWM captures how the exogenous physical environment evolves, the CDWM addresses the complementary question: how do the internal state of the network, queues, loads, slice allocations, protocol variables, and the resulting KPIs evolve over time under operator actions and the environmental context provided by the FWM? The CDWM is the action-conditioned core of the TWM. It takes as input a candidate control action, or a sequence of actions over a planning horizon, with spatial and contextual summaries produced by the FWM, and outputs predicted trajectories of KPIs and internal state variables. This capability allows the TWM to evaluate multiple candidate policies in a learned model before deploying any action in the live network. Conceptually, the CDWM combines ideas from latent-dynamics world models, model-based wireless control, and learned network surrogates~\cite{hafner2023dreamerv3,feriani2021wm_wireless,ferriol2022routenet,mobiworld2025}.

At this level, the CDWM operates at timescales from seconds to hours, bridging the gap between the fast physical-layer dynamics handled by the FWM and the session-level reasoning performed by TelecomGPT. Its outputs, KPI trajectories, predicted constraint violations, and uncertainty estimates, are the primary inputs on which TelecomGPT performs its planning, comparison, and guardrail enforcement. In this sense, the CDWM plays the role of the dynamics engine of the TWM, acting as the transition model in model-based reinforcement learning. 

\subsection{Representative Tasks of the CDWM}
\label{sec:cdwm_tasks}

\begin{table*}[t]
    \centering
    \begin{tabular}{p{2.8cm}p{4cm}p{10cm}}
        \toprule
        \textbf{Category} & \textbf{Task} & \textbf{Description} \\
        \midrule
        \multirow{3}{*}{KPI Prediction} &
        Throughput \& Latency Forecasting &
        Predict per-slice throughput, delay percentiles, and loss rates over a planning horizon. \\
        &
        Queue \& Buffer Dynamics &
        Model transient queue occupancy and overflow risk under candidate scheduling policies. \\
        &
        Resource Utilization Tracking &
        Forecast Physical Resource Block (PRB) and transport link utilization to detect congestion before it occurs. \\
        \midrule
        \multirow{2}{*}{Mobility \& Handover} &
        Handover Outcome Prediction &
        Predict handover success rates, ping-pong frequencies, and service interruptions under parameter changes. \\
        &
        Load Redistribution &
        Forecast how load-balancing policies shift traffic across cells over time. \\
        \midrule
        \multirow{2}{*}{Slicing \& Admission} &
        Per-Slice QoS Trajectory &
        Predict coupled QoS evolution across slices under joint resource allocation decisions. \\
        &
        Admission Impact Assessment &
        Evaluate how admitting a new tenant or slice affects existing SLA margins. \\
        \midrule
        \multirow{2}{*}{Energy \& Resilience} &
        Energy-Savings Impact &
        Model delayed coverage and latency effects of cell sleep, carrier shutdown, or power reduction. \\
        &
        Failure Propagation &
        Predict cascading effects of equipment failures or link outages across the network. \\
        \midrule
        \multirow{2}{*}{Counterfactual Evaluation} &
        Policy Comparison &
        Roll out KPI trajectories under alternative policies to estimate regret and rank candidates. \\
        &
        What-If Analysis &
        Answer counterfactual queries without executing actions on the live network. \\
        \bottomrule
    \end{tabular}
    \caption{Representative Tasks of the CDWM}
    \label{tab:cdwmlayer}
\end{table*}

In this section, we highlight the five different categories we envision for the representative tasks of the CDWM.

\textbf{KPI prediction:}
The most fundamental task of the CDWM is to predict how KPIs (throughput, latency, packet loss, and resource utilization) evolve over a specified planning window, and under a given configuration or policy. Unlike a static KPI predictor that maps current measurements to a point estimate, the CDWM produces {trajectories}, including sequences of predicted KPI values conditioned on the action sequence and the provided environmental context by the FWM. This temporal structure is important because many telecom effects are not immediate, for example, a change in scheduling policy may take minutes to affect the queue occupancy, while energy-saving actions can gradually shift load to neighboring cells. By capturing these dynamics, the CDWM enables the planner to reason about transient costs, not only steady-state outcomes.

\textbf{Mobility and handover:}
Handover and load-balancing decisions interact with user mobility in complex ways. The CDWM models how changes in handover parameters (for example, hysteresis, time-to-trigger, or cell individual offsets) affect handover success rates, ping-pong frequencies, and service interruption durations. At a longer timescale, it predicts how load-balancing policies redistribute traffic across cells, enabling the planner to anticipate congestion shifts rather than reacting to them after they occur.

\textbf{Slicing and admission control:}
In multi-tenant networks, the CDWM predicts how resource allocation and admission decisions at the slice level affect per-slice QoS trajectories. This includes modeling the coupling between slices, for example, allocating more resources to a URLLC slice may degrade eMBB throughput, and admitting a new tenant may push an existing slice towards its SLA boundary. Hence, the CDWM must maintain a representation of the multi-slice state and its coupled dynamics under joint control actions.

\textbf{Energy and resilience:}
Energy-saving actions such as cell sleep modes, carrier shutdown, and power reduction are inherently temporal. In specific, although they might reduce cost immediately, but might lead to a degraded coverage or increased latency with a delay that depends on traffic patterns and mobility. Similarly, equipment failures or link outages trigger cascading effects, rerouting, overload, and secondary failures, that unfold over minutes to hours. The CDWM models these propagation dynamics, allowing the planner to assess the full temporal cost of energy-saving strategies and the implications of failure scenarios.

\textbf{Counterfactual evaluation:}
The capability that most clearly distinguishes the CDWM from a conventional KPI predictor is \emph{counterfactual reasoning}, meaning the ability to answer ``what would have happened under a different policy?''  Given a candidate action sequence, the CDWM predicts the resulting state trajectory in its internal model without executing the action on the live network.  By comparing different predictions under multiple candidate policies, the planner can estimate regret, rank alternatives, and identify the policy that best satisfies the operator's objectives and constraints. This is the core mechanism through which the TWM provides foresight, instead of operating as a reactor only. 

The representative tasks of the CDWM are summarized in Table~\ref{tab:cdwmlayer}.

\subsection{Key Requirements of the CDWM}
\label{sec:cdwm_req}

The CDWM faces a distinct set of challenges compared to the FWM.  While the FWM must deal with the physics of propagation and the real-time constraints of the physical layer, the CDWM must learn dynamics that are driven by protocols, policies, and multi-agent interactions, mostly from limited and biased observational data.

\textbf{Action-conditioned causality:}
The CDWM must predict the effect of actions, in addition to the correlations between the configurations and KPIs. In networks, the observed data is generated under a specific behavioral policy, where the operator applies certain configurations in response to certain conditions. A model trained on such data learns to predict KPIs given the configurations that were historically chosen, which is fundamentally different from predicting the KPIs that will result from a novel configuration. This distinction, between prediction under the observed policy and prediction under interventions, is the core challenge of causal inference and offline reinforcement learning. Addressing this challenge requires techniques, such as importance weighting, doubly robust estimation, or explicitly modeling the logging policy, so that the CDWM can generalize to actions that differ from those in the training data.

\textbf{Multi-tenant and multi-timescale structure:}
Modern networks serve multiple slices, tenants, and service classes simultaneously, and the CDWM must represent their coupled dynamics.  An action that benefits one slice may harm another, also a fast scheduling change may interact with a slow energy-management policy in unexpected ways. Therefore, the CDWM must maintain a structured, multi-dimensional state that captures per-slice and per-cell variables, and it must learn the dynamics that span multiple timescales, from sub-second scheduling effects to hourly traffic patterns. Hierarchical or multi-rate architectures, in which faster dynamics are modeled at a smaller temporal resolution and slower dynamics at a coarser one, are a natural approach to this challenge.

\textbf{Sample efficiency and safe evaluation}
Training data for the CDWM is expensive, each data point corresponds to a real operational decision and its consequences, or to a costly simulation run.  The CDWM must accordingly be sample-efficient, extracting maximum information from limited trajectories. Furthermore, evaluating a learned CDWM is a safety-critical task. In specific, if the model is used to recommend actions that are then applied to a live network, errors in the model can lead to SLA violations or service degradation. This calls for careful evaluation protocols, including off-policy evaluation with confidence bounds, shadow-mode deployment where recommendations are logged but not executed, and gradual deployment where the model is given increasing autonomy as trust builds.  

\section{TelecomGPT / Foundation-Model Layer}
\label{sec:telecomgpt}

The FWM and CDWM provide the predictive mechanisms of the TWM, but they do not know what the operator wants or what constraints must be respected. The third layer, realized by a telecom-specialized foundation model or LLMs, which we refer to collectively as {TelecomGPT}, serves as the cognitive interface that connects the world models to the operator, external tools, and the network management systems. It extends recent telecom-specific LLM efforts from language assistance and tool use toward model-grounded planning and guardrail enforcement~\cite{zou2025telecomgpt,netgpt2024,wirelessllm2024,chatnet2023}.

TelecomGPT performs three complementary functions. First, it compiles high-level operator intents into the formal objectives and constraints that the FWM and CDWM require as inputs.  Second, it orchestrates calls to the world models, to external simulators and DTs, and to management plane tools, coordinating a multi-step planning process.  Third, it enforces safety constraints, non-negotiable SLA requirements, and policies, and produces human-readable explanations of its recommendations. In this way, TelecomGPT transforms the raw predictive capability of the TWM into actionable, auditable, and operator-aligned intelligence.

\subsection{Intent Translation}
\label{sec:tgpt_intent}

Operator requirements are typically expressed at a high level of abstraction, for example, ``ensure URLLC latency below 5\,ms during the stadium event'', ``reduce energy cost by 15\% without violating coverage SLAs'', or ``prepare the network for a planned maintenance window on cells 12--18''. TelecomGPT parses such intents and produces a machine-readable specification consisting of an objective function over CDWM-predicted KPI trajectories, hard constraints that candidate actions must satisfy, the action space to search over, and contextual information that should condition the FWM and CDWM. This compilation step allows the TWM to serve diverse operator needs without retraining the underlying world models. When an intent is under-specified, over-constrained, or contradictory, TelecomGPT detects these issues and interact with the operator to clarify, to avoid producing an infeasible plan.

\subsection{Tool Orchestration}
\label{sec:tgpt_tools}

Resolving a single intent typically requires multiple steps, including querying network state, generating candidate actions, scoring them through the FWM and CDWM, validating the top candidates through a DT, and presenting a ranked recommendation.  TelecomGPT orchestrates this pipeline by planning calls to the world models and, where needed, to external tools such as DT simulators and optimization engines. The orchestration logic may follow a fixed pipeline or a flexible agentic pattern in which TelecomGPT decides step by step which tool to call next. In either case, a structured planning trace is maintained for auditability and reproducibility.

\subsection{Guardrails, Safety, and Explainability}
\label{sec:tgpt_guardrails}

Given the fact that TelecomGPT lies between world-model predictions and action execution on the live network, it is the optimum candidate for safety enforcement. Before execution, TelecomGPT validates actions against SLA, regulatory, and safety constraints, combining CDWM predictions with runtime monitoring and recovery mechanisms. It also generates explanations that connect each recommendation to the underlying world-model outputs, helping operators to understand both what is recommended and why. In early deployments, TelecomGPT operates in advisory mode, requiring explicit operator approval, then, as trust accumulates through validated prediction accuracy, the system can be granted more autonomy for low-risk actions, while keeping the human-in-the-loop approval for novel or high-impact decisions.

In practice, realizing TelecomGPT involves determining different factors, including the interaction pattern (prompt-based, function-calling APIs, or full agentic frameworks), deployment location (cloud, on-premises, or edge), and degree of domain specialization (general-purpose LLMs with telecom prompting versus a model fine-tuned on telecom corpora). These choices are not mutually exclusive and may be combined depending on the operator context.

\section{What TWM Adds Beyond Existing Paradigms}
\label{sec:comparison}

Building on the TWM architecture presented above, we now compare TWMs with similar paradigms over different operational capabilities. Table~\ref{tab:twm_comparison} compares LLM-based agents, standalone DTs, existing world-model paradigms, and the proposed TWM, with respect to six core capabilities. In what follows, we analyze each capability in detail, explaining why current paradigms are not sufficient, and how the TWM architecture addresses the gap. 
\begin{table*}[t]
\centering
\footnotesize
\renewcommand{\arraystretch}{1.05}
\setlength{\tabcolsep}{3pt}
\begin{tabular}{lcccccc}
\toprule
\textbf{Paradigm} &
\textbf{Telecom state} &
\textbf{Action-conditioned rollouts} &
\textbf{Explicit uncertainty} &
\textbf{Multi-timescale dynamics} &
\textbf{Policy planning} &
\textbf{LLM + guardrails} \\
\midrule
LLM-based agents
& \pmark  & \xmark & \xmark & \xmark & \pmark & \cmark \\
Digital Twins
& \cmark  & \pmark & \pmark & \cmark & \pmark & \xmark \\
World models
& \pmark  & \cmark & \pmark & \pmark & \cmark & \xmark \\
\addlinespace[0.15em]
\textbf{TWM}
& \textbf{\cmark} &
  \textbf{\cmark} &
  \textbf{\cmark} &
  \textbf{\cmark} &
  \textbf{\cmark} &
  \textbf{\cmark} \\
\bottomrule
\end{tabular}
\caption{Qualitative comparison of LLM-based agents, Digital Twins, world-model paradigms, and the proposed TWM across core capabilities.  Symbols reflect the typical implementations: native (\cmark), partially present (\pmark), or not typically provided (\xmark).  Individual systems may differ.}
\label{tab:twm_comparison}
\end{table*}

\subsection{Telecom State Grounding}
LLM-based agents process KPI summaries and alarm logs as token sequences, however, it does not maintain a structured, machine-readable network state that links throughput to PRB allocations, interference graphs, or mobility patterns. DTs maintain rich simulator internal state, but it highly relies on the implementation and generally not exposed for external learning or planning algorithms. Generic WMs learn structured latent states grounded in image pixels or proprioceptive readings, however, not in telecom quantities such as spectral efficiency maps, queue occupancies, or slice-level SLA metrics. The TWM addresses this gap through its two-world factorization, where the CSW captures operator-actuated network state while the EGW captures propagation, traffic, mobility, and failures. These models provide a shared, queryable state foundation that all three TWM layers can read from and write to.


\subsection{Action-Conditioned Predictions}
A defining WM property is the ability to answer ``what if'' questions. In specific, given the current state and a candidate action sequence, the WM to be able to predict the resulting future trajectory.  LLM-based agents can reason verbally about consequences, but cannot simulate transient dynamics of a scheduler change or cascading effects of a link failure through queuing and routing layers.  DTs can perform such rollouts, but each requires a full simulation run taking seconds to minutes, making exhaustive search over combinatorial action spaces prohibitive.  Generic WMs provide fast learned rollouts, but their state and action spaces target games or robotics, not telecom control actions, such as PRB partitioning vectors or slice admission thresholds.  The TWM's CDWM layer is designed as a telecom-native learned dynamics model that takes structured telecom state and a control action as input and produces predicted next state with KPI outcomes within the CSW/EGW. Within this context, CDWM rollouts are significantly faster than DT runs, enabling model-based search over large action spaces.


\subsection{Explicit Uncertainty}
Telecom planning under uncertainty, including sudden increase in demand, equipment failures, and interference from neighboring cells, requires calibrated confidence intervals rather than point predictions alone. LLM-based agents express uncertainty in natural language, for example, ``this change might cause congestion'', which is neither quantitative nor actionable for automated decision loops. DTs can quantify uncertainty through Monte Carlo simulation, but running many replicas radically increases simulation cost. Generic WMs incorporate stochastic latent variables, e.g., the RSSM stochastic state~\cite{hafner2023dreamerv3}, but these are typically calibrated for perceptual prediction error rather than telecom risk metrics such as SLA violation probability or tail latency. In the TWM, uncertainty is propagated through the predictive layers, where probabilistic FWM outputs provide spatial confidence bounds, while the CDWM produces prediction intervals for KPIs such as per-slice throughput, delay percentiles, and queue loss rates. These estimates then feed into the planning and guardrail logic of the TWM, where candidate actions whose predicted violation probability exceeds a threshold can be rejected or down-weighted before being surfaced to the operator, thereby enabling risk-aware decision-making.

\subsection{Multi-Timescale Dynamics}
Telecom networks evolve across multiple timescales, from millisecond fading and scheduling to mobility, traffic variations, and longer-term capacity planning.
LLM-based agents do not enjoy an explicit temporal model. In particular, they process snapshots without capturing the causal chain linking a scheduling decision to a congestion event minutes later.  DTs can in principle model multiple timescales, but most platforms operate at a single simulation resolution, and bridging sub-millisecond PHY simulation to hour-scale planning within one run is computationally intractable. Finally, generic WMs similarly operate at a single fixed time step.  The TWM's layered architecture is explicitly designed to span over these timescales, in which the FWM captures fast physical-layer dynamics, such as channel coherence, beam tracking and interference, the CDWM models control-loop dynamics at RAN and transport timescales from a slot-level to a minute-level, and TelecomGPT reasons at the strategic timescale. Hierarchical rollouts allow the CDWM to abstract over FWM predictions while TelecomGPT abstracts over CDWM trajectories, yielding a tractable multi-resolution planning framework.


\subsection{Policy Planning}
The ultimate purpose of a predictive model in a network automation stack is to support planning, including selecting a policy or action sequence that optimizes an objective subject to constraints. LLM-based agents plan heuristically through chain-of-thought prompting, driven by pattern matching rather than systematic search through a model of consequences.  DTs support scenario-based evaluation but provide no search mechanism, and hence, the burden of proposing and iterating over candidates falls on an external optimizer. Generic WMs integrate tightly with planning algorithms such as model-predictive control or tree search~\cite{schrittwieser2020muzero}, but their action spaces have no relation to telecom control variables. The TWM combines model-based planning with telecom-native action spaces, where the CDWM's differentiable dynamics enable gradient-based optimization, while its fast rollouts support sampling-based search, e.g., cross entropy method or Monte Carlo tree search, over discrete or hybrid action spaces such as slice resource vectors. Crucially, the planning objective can incorporate hard constraints, like SLA targets, regulatory limits, and energy budgets, as penalties or barriers, ensuring that the search respects operational requirements that a generic planner would be unaware of.


\subsection{LLM Integration and Guardrails}
Neither DTs nor generic WMs provide a natural-language interface for operator interaction or safety guardrails that prevent execution of harmful actions. DTs simulate any configuration, including constraint-violating ones, and generic WMs are similarly agnostic to domain-specific safety requirements. LLM-based agents excel at natural-language interaction and can be equipped with output filters or tool-use permissions, but they lack the predictive dynamics needed to assess whether a proposed action is safe in practice, as opposed to safe in principle.  The TWM couples a TelecomGPT / foundation-model layer with world-model predictions and constraint enforcement. When an operator issues a high-level intent such as ``prepare for a 30\% traffic surge in the downtown macro cells,'' TelecomGPT translates this into candidate CDWM-level action sequences, the CDWM rolls out their consequences, and a guardrail module verifies that predicted KPIs remain within SLA bounds before any action reaches the live network.  This closed-loop architecture ensures that natural-language accessibility does not come at the cost of operational safety.


\section{Building and Evaluating TWMs in Practice}
\label{sec:building}

The previous sections defined what a TWM is and what it should do.  This section turns to the practical question of how to build a TWM, including what data is needed, how each layer should be trained, and how the resulting system should be evaluated.

\subsection{Data Pipelines}
\label{sec:build_data}

\begin{figure}[t]
\centering
\includegraphics[width=\columnwidth]{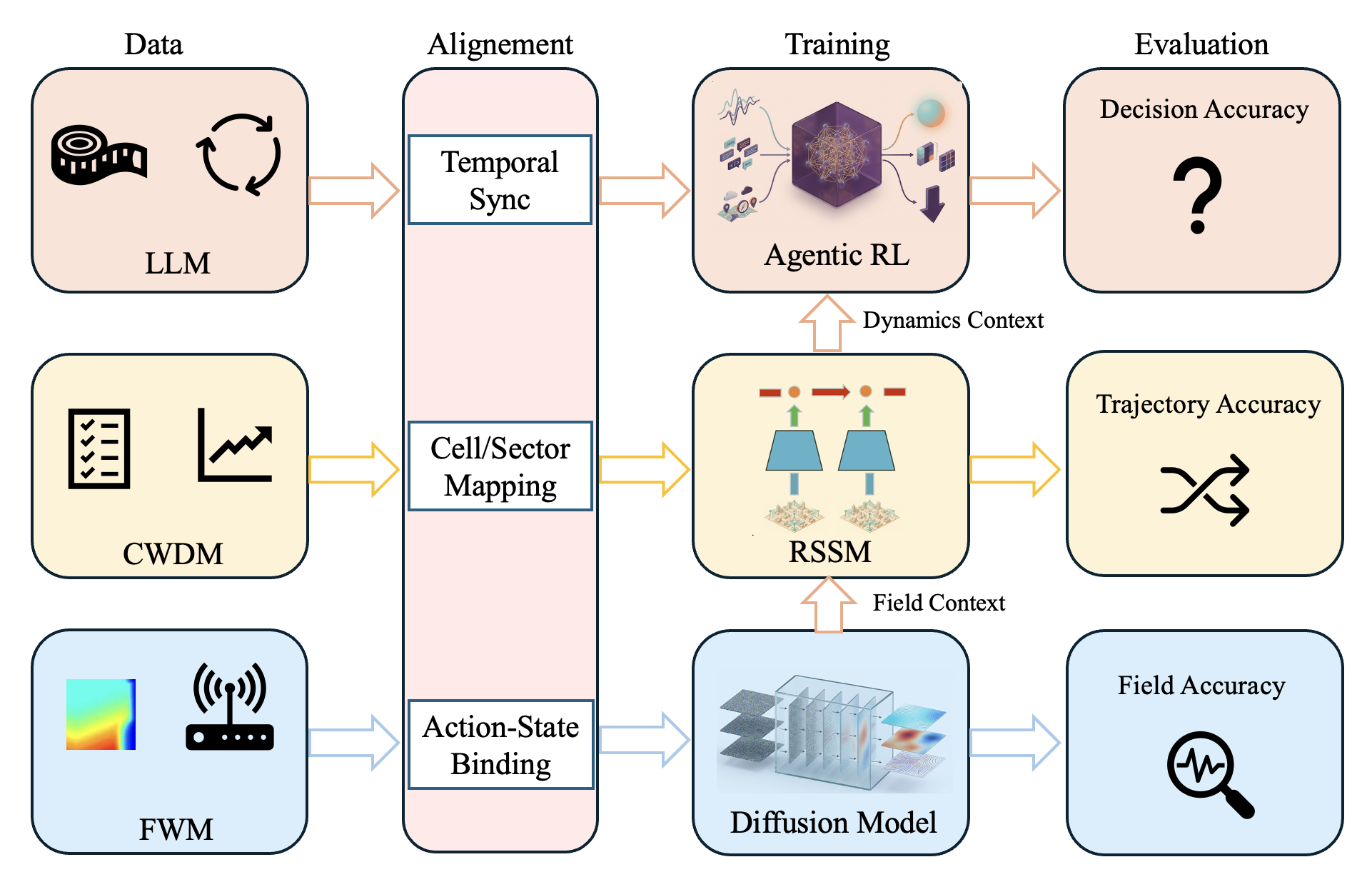}
\caption{Layer-specific data, alignment, training, and evaluation pipeline for TWM. Physical observations, network trajectories, and semantic telecom knowledge are aligned through temporal synchronization, cell/sector mapping, and action-state binding before being used to train the FWM, CDWM, and TelecomGPT layer; cross-layer field and dynamics context are then propagated upward, and each layer is evaluated through field, trajectory, or decision accuracy.}
\label{fig:data_ecosystem}
\end{figure}

Each TWM layer relies on a distinct data stream, and a cross-layer alignment step is required to integrate them, as illustrated in Fig.~\ref{fig:data_ecosystem}.

\textbf{FWM Data}
The FWM relies on geo-referenced radio-frequency observations and rich environmental context to capture the interaction between wireless signals and the physical world. Data sources include both real measurements and model-generated information. Measurement data provide spatial signal observations, while environmental information characterizes the surrounding geometry, terrain, and propagation conditions. 

In addition, channel state information (CSI) across time, frequency, and space serves as a fundamental input, offering detailed descriptions of channel variations and enabling the model to learn dynamic propagation behaviors. To complement real-world data, simulation data generated by high-fidelity NDTs can produce dense and well-structured signal representations under controlled settings. These synthetic data sources are particularly important for pre-training, allowing the FWM to learn generalizable physical patterns even in scenarios with limited measurements.

\textbf{CDWM Data}
The CDWM requires trajectories, including time-ordered sequences of (state, action, next-state, KPI) tuples, that capture how the network evolves under control decisions. On the live-network side, relevant sources include RAN and core performance counters (per-cell throughput, latency percentiles, PRB utilisation), O-RAN E2 interface telemetry, OSS/BSS configuration change logs that record which parameters were modified and when, and alarm and event streams. In O-RAN-like deployments, E2 telemetry, A1 policies, and xApp/rApp traces provide a key source of supervisory data for learning and validating the CDWM. On the simulation side, DT-generated synthetic trajectories are critical for two reasons, first, they allow controlled injection of rare events (mass failures, extreme load spikes, correlated outages) that would be dangerous or impractical to trigger in a live network, and they provide ground-truth alternative outcomes for the same scenario under different policies, which are not available from operational logs. More importantly, the {logging policy} under which each trajectory was generated must be recorded alongside the data.  As discussed in Section~\ref{sec:cdwm_req}, the CDWM must learn the causal effect of actions, not only correlations under the historical policy, as without knowledge of the logging policy, off-policy correction techniques cannot be applied.

\textbf{TelecomGPT Data}
The data requirements for the TelecomGPT layer are largely addressed by existing telecom foundation-model frameworks~\cite{zou2025telecomgpt}, comprising large-scale pre-training on 3GPP standards, protocol specifications, configuration documentation, and troubleshooting corpora, followed by supervised fine-tuning on domain tasks. Also, a corpus of structured intent-to-specification examples that map operator intents (natural-language or formal) to the machine-readable objectives, constraints, and action spaces that drive the FWM and CDWM planning loop, are essential for the TelecomGPT layer.

\textbf{Cross-Layer Alignment and Governance}
Fig.~\ref{fig:data_ecosystem} summarizes the end-to-end data flow from heterogeneous telecom sources to layer-specific training signals.
The interaction patterns described in Section~\ref{sec:twm} require the FWM's spatial outputs and the CDWM's temporal state observations to be aligned in both time and space.  In practice, this means that RF measurement timestamps must be synchronized with counter reporting intervals, and that spatial fields must be aggregated to the cell or sector granularity at which the CDWM operates. Data governance is also a concern, as operational network data is sensitive and subject to regulatory constraints on storage and processing.  Privacy-preserving techniques (differential privacy, federated learning) and synthetic-data-first approaches, in which models are pre-trained primarily on DT-generated data and adapted with minimal real data, can help address these constraints.

\subsection{Training and Alignment of TWM Components}
\label{sec:build_train}

We adopt a modular, layer-by-layer training strategy in which each TWM component is trained with the approach best suited to its learning problem, followed by interface alignment across layers.

\textbf{FWM Training:}
Among the world-model paradigms surveyed in Section~\ref{sec:bg_wm}, the generative family, and specifically {conditional diffusion models}, is the most promising candidate for the FWM.  Just as Sora~\cite{sora2024} and UniSim~\cite{yang2024unisim} learn to generate future video frames conditioned on actions, a diffusion-based FWM can learn to generate future spatial RF fields conditioned on infrastructure configuration and environmental context.  

A practical FWM can first aggregate heterogeneous channel observations across time, space, and frequency into a compact latent representation, which then serves as a conditioning signal for the diffusion model together with coarse channel estimates and configuration vectors. Action-conditioning can be introduced through cross-attention, enabling prediction of wireless fields across time, frequency, and space while maintaining global consistency. In this way, the diffusion model refines coarse observations into high-fidelity channel or field realizations, which can support downstream tasks such as channel estimation, signal detection, and resource optimization. Moreover, multiple samples from the conditional distribution naturally provide uncertainty estimates and help address the partial-observation challenge highlighted in Section~\ref{sec:fwm}.

Recent work has begun to explore this direction for wireless channel estimation and related inference tasks. In particular, channel reconstruction can be formulated as a Bayesian inverse problem, where diffusion models learn implicit channel priors and enable posterior sampling conditioned on pilots and noise statistics~\cite{9957135,10930691}. More recent studies further incorporate structured side information and signal-processing modules to improve generalization to non-stationary channels and complex environments~\cite{11363381, yang2025diffusionmodelswirelesstransceivers}. A practical challenge is the trade-off between model expressiveness and real-time inference, which motivates a two-tier design, a full diffusion model for offline, planning-timescale field generation, and a distilled lightweight model for real-time FWM inference at the channel-coherence timescale. In both tiers, physics-informed priors from 3GPP channel models and site-specific geometry can regularize training and improve robustness to unseen deployments.

\textbf{CDWM Training:}
The latent dynamics paradigm, represented by the Dreamer/RSSM family~\cite{hafner2019planet, hafner2020dreamer, hafner2023dreamerv3}, is a natural fit, given the fact that the CDWM's core task is action-conditioned KPI rollout. The transition model is built for exactly this purpose, and its stochastic and deterministic state components help separate predictable control effects from stochastic traffic and interference variability.

Training the CDWM for telecom requires several adaptations. The image-based encoder must be replaced with a structured telecom encoder that ingests configuration vectors, KPI time series, graph-structured topology, and FWM-derived spatial summaries, with the latent state representing multiple slices and tenants jointly to capture inter-slice resource competition. A hierarchical or multi-rate RSSM, with fast state components for sub-second scheduling effects and slow components for hourly traffic patterns, is needed to cover the multi-timescale dynamics illustrated in Fig.~\ref{fig:timescales}.

Off-policy correction techniques (importance weighting, doubly robust estimation, or conservative value estimation) must be incorporated, since operational trajectories are collected under specific logging policies and the CDWM must generalize to novel actions not seen during training.  As an additional design choice, ensemble methods over multiple RSSM instances can quantify model uncertainty, providing prediction intervals that the TelecomGPT layer can use for risk-aware planning.

\textbf{TelecomGPT: Interface Alignment}
The TelecomGPT layer does not require training a foundation model from scratch. Instead, frameworks such as TelecomGPT~\cite{zou2025telecomgpt} already provide telecom-specialized LLMs. For this layer, the TWM interface alignment, where the FWM and CDWM must be exposed as callable tools, and TelecomGPT must be able to reason over their structured predictive outputs (KPI trajectories, uncertainty intervals, violation probabilities). Additionally, guardrail enforcement must be grounded in model-based uncertainty instead heuristic rules.  Targeted fine-tuning on these capabilities enables the intent compilation and risk-aware orchestration described in Section~\ref{sec:telecomgpt}.

\textbf{Training Order and Iteration}
We adopt a modular, bottom-up training order, starting by the FWM on spatial field data, then CDWM conditioned on frozen FWM outputs, and finally TelecomGPT fine-tuned against the trained world-model back-ends.  This sequence can be iterated as each layer improves, with discrepancies between CDWM predictions and DT or live-network outcomes guiding targeted retraining and tuning. Continual adaptation is essential for tracking non-stationary network conditions, but must be paired with shadow-mode deployment to prevent model drift from degrading operational decisions.

\subsection{Towards TelecomWorldBench}
\label{sec:build_eval}

Currently, there are no established benchmarks available for evaluating TWMs end to end. In this section, we outline here the vision for benchmark and evaluating TWMs, which we tentatively call \textbf{TelecomWorldBench}.  As illustrated in the Evaluation column of Fig.~\ref{fig:data_ecosystem}, each TWM layer requires its own evaluation axis.

\textbf{Field Accuracy (FWM)}
The FWM should be assessed on field reconstruction error, e.g., Root Mean Square Error (RMSE) and structural similarity, against ray-tracing or ground truth measurement, calibration of uncertainty intervals (coverage probability at nominal confidence levels), and spatial generalization to unseen urban morphologies or deployment regions.

\textbf{Trajectory Accuracy (CDWM)}
The CDWM should be assessed on KPI trajectory accuracy (per-step and cumulative error), calibration of predicted violation probabilities against observed outcomes, and the relationship between ensemble disagreement and true prediction error. Calibration is particularly important. In specific, a CDWM whose predicted 95\% confidence interval covers the true outcome only 70\% of the time will produce overconfident plans that underestimate risk.

\textbf{Decision Accuracy (TelecomGPT)}
Prediction quality alone is insufficient, but also whether the full TWM enables better decisions. Decision-accuracy evaluation measures the end-to-end utility of the planning pipeline. Key metrics include aggregate utility, SLA pass rate, constraint satisfaction margins, and regret relative to the optimality.  

All three axes should additionally be tested for robustness, including the evaluation of TWM's capabilities in out-of-distribution generalization, rare-event handling, and distribution-shift tracking over time.


\section{Unique Use Cases Enabled by TWM}
\label{sec:usecases}

This section highlights potential planning scenarios in which the full TWM stack is needed. The goal is not to provide an exhaustive application roadmap, but to identify classes of decision problems that require the joint use of (i) field-level forecasting of the exogenous environment, (ii) action-conditioned rollout of network-state and KPI trajectories, and (iii) intent- and constraint-aware orchestration. In each scenario, the key question is not if the network can be analyzed, but whether alternative futures can be compared efficiently and safely before any action is applied to the live system.

\subsection{Intent-Grounded RAN and Slicing Reconfiguration}
Consider an operator intent, such as ``prioritize URLLC reliability in critical zones during evening demand peaks, while keeping total resource cost near the daytime baseline.'' This is not a single-parameter tuning problem, but a constrained planning task over multiple candidate actions, such as slice resource reallocation, scheduler-weight adjustment, mobility-parameter updates, and selective cell reconfiguration. The difficulty is that the operator's requirement is expressed at a high semantic level, while the operational impact of each candidate action depends on both current network conditions and how the radio environment is expected to evolve over the planning horizon.

In this setting, the FWM first forecasts where service risk is likely to emerge in space, for example through predicted coverage degradation, mobility-driven hotspot formation, or interference concentration in critical zones. These field-level predictions provide the contextual substrate on which the CDWM evaluates candidate macro-actions and rolls out their consequences in terms of queue evolution, slice-level resource usage, end-to-end latency, reliability, and SLA-violation risk. TelecomGPT compiles the operator's intent into an explicit optimization problem, including objective weights, admissible action sets, and hard guardrails, then orchestrates FWM and CDWM queries to compare alternative futures and produce an auditable recommendation.  The TWM uniquely combines semantic intent understanding, spatial awareness, and temporal rollout into a single planning loop.


\subsection{Robust Contingency Planning Under Exogenous Shocks}
A second use case arises before predictable but uncertain stress windows, such as stadium events, public gatherings, severe weather conditions, or planned maintenance operations. In such situations, the operator must choose among contingency policies before the event unfolds. For example, pre-activating dormant cells, reserving additional resources for critical slices, tightening or relaxing mobility parameters, modifying admission thresholds, or rerouting transport capacity. The challenge is that the dominant uncertainty is external and uncontrolled. Crowd density, blockage patterns, weather-induced propagation degradation, and correlated load surges are not directly controllable, yet they strongly determine whether a contingency policy will succeed or fail.

In this setting, the FWM generates scenario-conditioned forecasts of the evolving field environment, such as coverage-risk maps, interference hotspots, mobility corridors, and spatial load concentration under different shock hypotheses. The CDWM then rolls out the resulting KPI trajectories under candidate contingency plans, producing risk-aware forecasts of slice-level throughput, latency, reliability, and violation probability over the planning horizon. TelecomGPT translates operator requirements, for example preserving URLLC reliability while minimizing temporary over-provisioning cost, into explicit objectives and risk thresholds, orchestrates scenario sweeps across candidate policies, and ranks the resulting plans before sending a small shortlist to a DT for final validation. The TWM thus enables robust pre-event planning that combines spatial awareness of the disturbance with fast policy screening under explicit uncertainty.


\subsection{Cross-Domain Planning Across RAN and Transport}
A third use case concerns decisions whose impact spans both the radio access network and the xHaul/transport domain. Consider a multi-slice deployment in which radio congestion in a hotspot region can be relieved by changing slice shares, mobility parameters, or cell operating points, but where these same actions alter traffic ingress patterns and may create new bottlenecks in midhaul or backhaul links. In such settings, domain-wise optimization is often insufficient, as a radio-side decision that improves cell-level KPIs may simultaneously increase transport queuing delay, while a transport-side rerouting decision may preserve link utilization at the expense of radio-layer reliability or handover stability.

The TWM is valuable here because it provides a shared predictive foundation across domains. The FWM forecasts where demand, coverage stress, and mobility pressure are likely to emerge spatially, thereby identifying the radio conditions that drive downstream transport load. Conditioned on this information, the CDWM rolls out the coupled evolution of slice allocations, queue occupancy, link utilization, and end-to-end KPIs under candidate actions such as slice-share reallocation, scheduler-policy changes, mobility-parameter tuning, queue-priority updates, or routing adjustments. TelecomGPT compiles the operator's intent into a cross-domain objective, for example preserving URLLC latency while limiting degradation to eMBB throughput and keeping transport utilization within safe margins, and then explains the resulting trade-offs in operator-facing language. Unlike disconnected tool chains, the TWM evaluates cross-domain alternatives as coherent counterfactual futures. The PoC in Section~\ref{sec:poc} represents this scenario.


\section{Proof-of-Concept Demonstration}
\label{sec:poc}

\subsection{Scenario and Control Problem}
We design the TWM architecture on a representative cross-domain network slicing problem spanning RAN and transport.
This PoC provides a validation on one scenario family, not the full TWM concept.

\begin{figure*}[t]
\centering
\subfloat[Spatial layout]{\includegraphics[width=0.85\columnwidth]{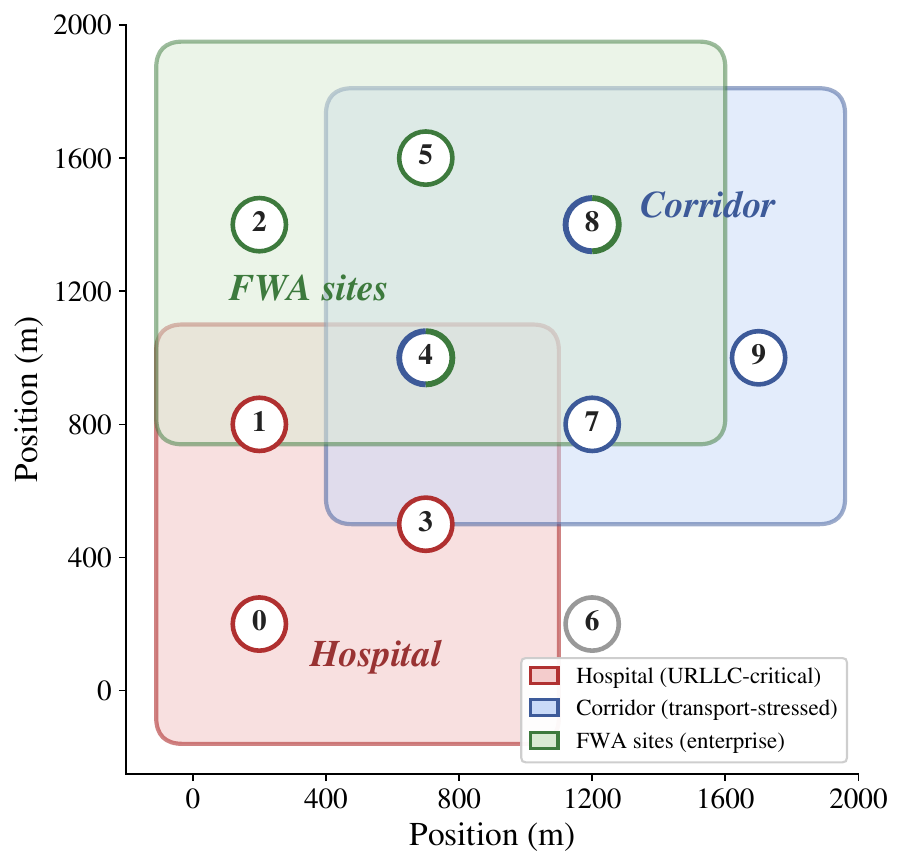}\label{fig:scenario_spatial}}%
\hfill
\subfloat[Transport topology]{\includegraphics[width=1.15\columnwidth]{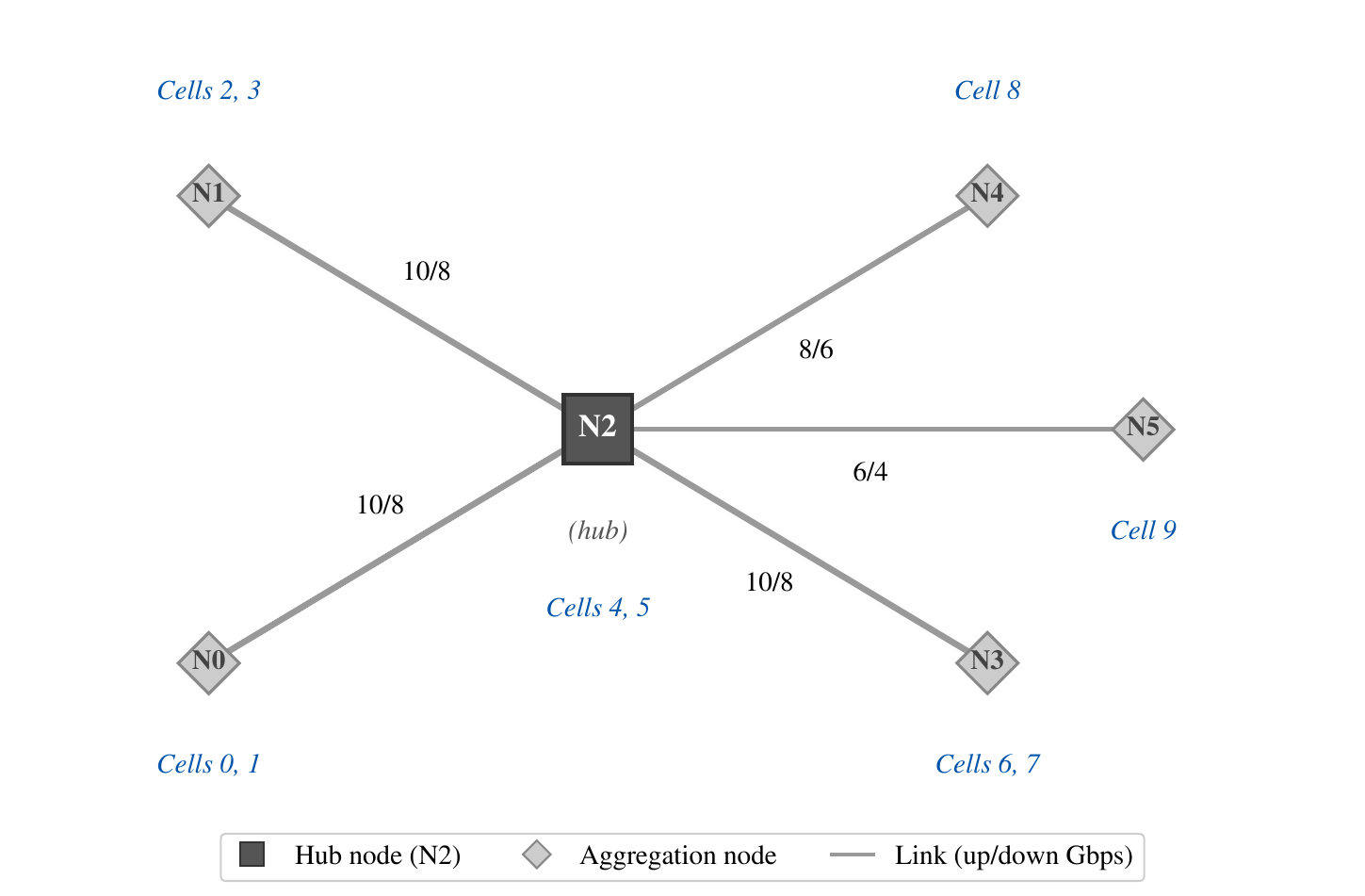}\label{fig:scenario_topo}}
\caption{PoC environment.  (a)~Shaded areas are schematic operational zones; colored cell outlines indicate region-associated cells.  Cells~4 and~8 belong to both corridor and FWA regions (split blue/green outlines).  Cell~6 lies outside all named regions.  (b)~Hub-and-spoke transport with 6~aggregation nodes (N0--N5) and 10~directed links.  Pairs of cells share aggregation nodes; link labels show uplink/downlink capacity in Gbps.}
\label{fig:scenario}
\end{figure*}

The environment models a 10-cell RAN deployment on a $2{\times}2$\,km grid serving three slices, eMBB (best-effort broadband), URLLC (latency-critical), and FWA (fixed wireless access). As shown in Fig.~\ref{fig:scenario}, three overlapping operational zones create the central planning tension, the hospital zone concentrates URLLC-critical demand, FWA enterprise sites create localized fixed-wireless load, and the corridor funnels uplink flows through shared transport links. Cells~4 and~8 belong to both corridor and FWA regions, so resource decisions in these cells affect two competing SLA constraints simultaneously. The network follows a hub-and-spoke topology with six aggregation nodes and ten directed links, connecting the radio and backhaul domains. When links fail, paths are disrupted and traffic is rerouted, so local radio decisions impact the broader network.

\textbf{Action Space}
At each slow-timescale step the planner decides how to share radio and transport resources among the three slices.
Formally, a slicing action consists of PRB share allocations $(p_{c,s}) \in [0,1]^{10 \times 3}$ across cells and slices, and transport BW share allocations $(b_{\ell,s}) \in [0,1]^{10 \times 3}$ across links and slices, subject to $\sum_s p_{c,s} \leq 1$ and $\sum_s b_{\ell,s} \leq 1$.
A policy is a sequence of 12~such actions over the planning horizon.

\textbf{Two-timescale Dynamics}
The planner operates at a slow timescale representing minutes-scale RIC control loops ($T_\mathrm{slow}{=}12$ steps per horizon).
Each slow step triggers $T_\mathrm{fast}{=}15$ fast-timescale queuing sub-steps that resolve per-cell scheduling, buffer dynamics, and transport congestion.
A planner that ignores the fast timescale cannot predict whether a seemingly efficient allocation will cause queue overflows and latency spikes.

\textbf{Exogenous Shocks}
Two families of demand shocks are used, namely, \emph{hotspot} (random spatial surges near the hospital zone) and \emph{wave} (corridor-propagating events that stress transport links).
In-distribution (ID) shocks use base-rate parameters, while out-of-distribution (OOD) shocks amplify intensity, duration, and failure probabilities, yielding four evaluation scenarios that test both interpolation and extrapolation.

\textbf{KPIs}
We report both compliance and efficiency.
Compliance is measured through per-slice SLA violation probabilities $\mathbb{P}_\mathrm{urllc}^\mathrm{V}$, $\mathbb{P}_\mathrm{fwa}^\mathrm{V}$, and a global indicator $\mathbb{P}_\mathrm{g}^\mathrm{V}$ that is triggered when any per-slice loss ceiling is exceeded (URLLC~\( 99 \%\) latency ${>}20$\,ms or loss ${>}10\%$; FWA loss ${>}20\%$; eMBB loss ${>}25\%$).
Efficiency is measured as cost reduction $\Delta_\mathrm{cost}$ relative to uniform $\frac{1}{3}$ allocation where cost is a normalized control cost that reflects how much resource is allocated and how rapidly the policy changes over time.

\subsection{TWM Implementation}

\textbf{FWM}
The Field World Model provides spatial awareness by predicting 12-dimensional field features (per-slice capacity, load, critical-zone stress, transport bottleneck, and SINR) from the grid of input channels.
It is realized as a 3-layer CNN trained via MSE regression on analytic-model outputs, achieving $R^2{=}0.99$ on held-out samples.

\textbf{CDWM}
The Control/Dynamics World Model predicts how KPIs evolve over time under a candidate policy.
It is realized as an RSSM ensemble ($N{=}3$ members, 96~deterministic $+$ 32~stochastic dimensions) trained with evidence lower bound (ELBO) loss on ID trajectories.

\textbf{TelecomGPT}
The LLM layer (Qwen2.5-7B-Instruct \cite{yang2024qwen2-5}) fulfills two roles,
(i)~prescreening (rejecting structurally unsafe candidates before surrogate scoring, at zero FWM/CDWM cost), and
(ii)~corner-case intervention (generating new resource allocations via parameterized templates when the planner detects high FWM violation risk or CDWM uncertainty).

\textbf{Planner}
The planner uses a two-phase receding-horizon architecture.
In Phase~1, FWM$+$CDWM surrogates score candidates and the top~3 are simulator-verified (12~sim calls per horizon, 384~per scenario over 32~horizons); if none pass, the planner falls back to uniform allocation. In Phase~2, the selected plan is executed step by step.  At each step the FWM and CDWM monitor for corner cases, and when triggered the LLM proposes alternative actions that must pass a do-no-harm utility gate before application.

\subsection{Evaluation Protocol and Baselines}

A library of 256~candidate policies is generated from the city configuration.
The TWM planner keeps the top 128, prescreen them to ${\sim}80$ via the LLM, surrogate-ranks the survivors, and simulator-verifies the top~3.
Phase~2 then executes the best-verified plan with selective LLM intervention.
All results report means over 32~horizon seeds per scenario, surrogate evaluations are not counted as simulator calls.
We now evaluate whether this integrated pipeline improves compliance and cost tradeoffs relative to the baselines.

\textbf{SLA Compliance Margin}
To measure the distance from SLA violation on a continuous scale, we define the \emph{SLA Compliance Margin} (SCM) as the minimum per-slice headroom before reaching the respective loss ceiling, as follows
\begin{equation}
\mathrm{SCM} = \min\!\left(
  1 {-} \tfrac{\ell_\mathrm{urllc}}{\tau_\mathrm{urllc}},\;
  1 {-} \tfrac{\ell_\mathrm{fwa}}{\tau_\mathrm{fwa}},\;
  1 {-} \tfrac{\ell_\mathrm{embb}}{\tau_\mathrm{embb}},\;
  1 {-} \tfrac{d_\mathrm{urllc}^{99}}{\tau_d}
\right)^{+}\!\!.
\label{eq:scm}
\end{equation}
Here, $\ell_s$ is the average loss rate of slice~$s$, $d_\mathrm{urllc}^{99}$ is the 99th-percentile URLLC latency, and the $\tau$ values denote the respective SLA ceilings ($\tau_\mathrm{urllc}{=}10\%$, $\tau_\mathrm{fwa}{=}20\%$, $\tau_\mathrm{embb}{=}25\%$, $\tau_d{=}20$\,ms).
$\mathrm{SCM}{=}0$ means at least one slice is at or beyond its limit, and higher values indicate more balanced headroom across slices.

\textbf{Baselines}
Four baselines are compared against the TWM:
\begin{itemize}
\item \textsc{DT-fair}: budget-matched DT search (384~sim calls), evaluating 24~candidates with 16~seeds each.
\item \textsc{FWM}: spatial scoring only (EGW), no learned dynamics.
\item \textsc{CDWM}: RSSM ranking only (CSW), no spatial features.
\item \textsc{LLM Agent}: LLM allocates resources without any world model.
\end{itemize}
\subsection{Results}

\begin{figure}[t]
\centering
\includegraphics[width=\columnwidth]{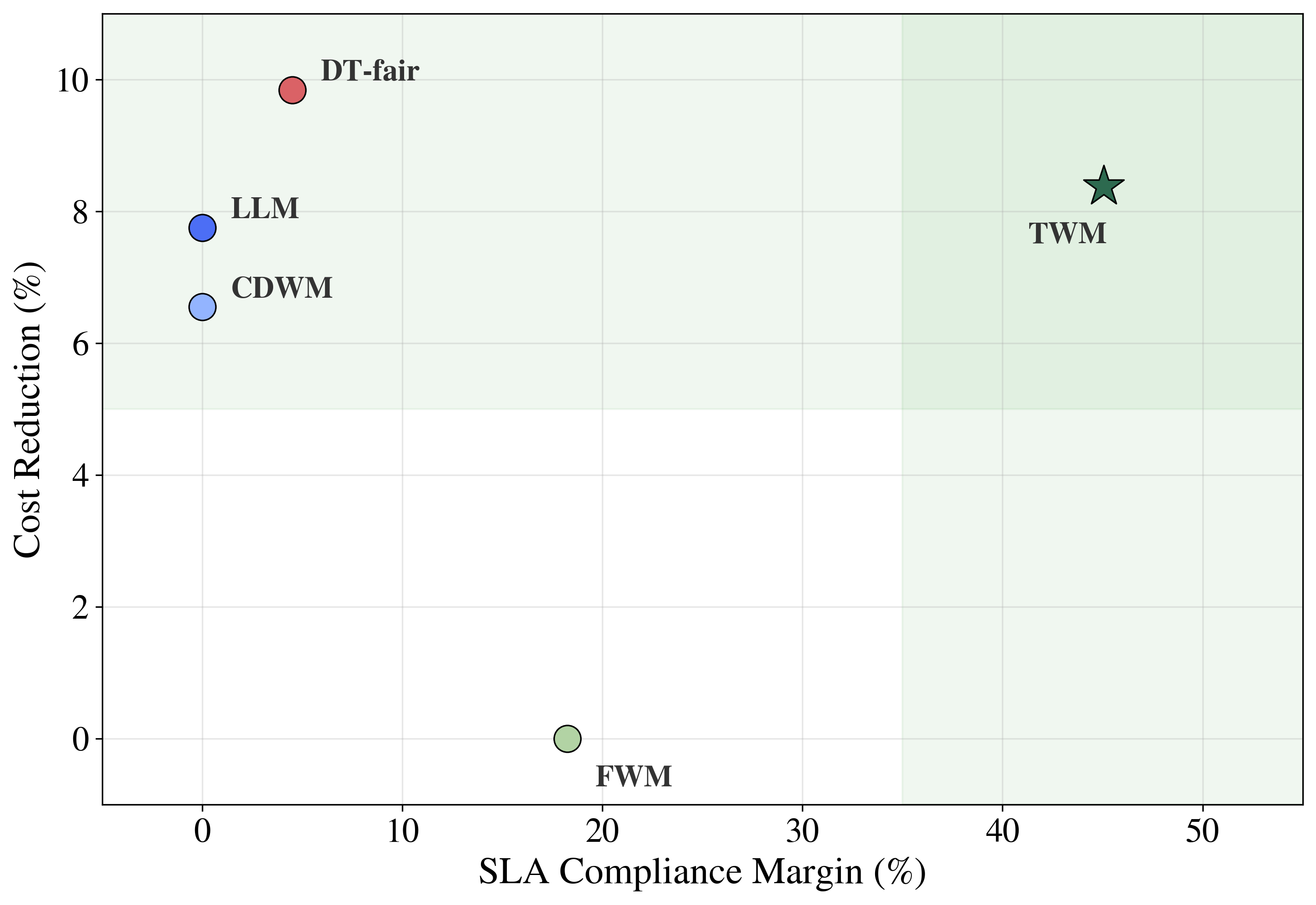}
\caption{Cost--compliance Pareto front averaged across four scenarios.  TWM (star) is the only method in the Pareto-optimal zone, with high SCM and cost savings. FWM achieves compliance but no savings; DT-fair and single-world baselines achieve savings but violate SLAs.}
\label{fig:pareto}
\end{figure}

Fig.~\ref{fig:pareto} places these outcomes on the cost--compliance Pareto front. TWM is the only method in the top-right region, combining average $\mathrm{SCM}{=}45\%$ with $\Delta_\mathrm{cost}{=}8\%$.
None of the single-layer baseline achieves both SLA compliance and cost savings. The LLM-only agent achieves $\mathrm{SCM}{=}0$ in all scenarios, without a world model it selects policies with 27\% URLLC loss and \( 99 \%\) latency above 300\,ms.
CDWM-only also reaches $\mathrm{SCM}{=}0$ because RSSM predictions without FWM spatial conditioning miss URLLC queue starvation.
FWM-only achieves moderate compliance (avg $\mathrm{SCM}{=}18\%$) through balanced spatial scoring, but it has no cost reduction.  Spatial features alone cannot optimize allocation without temporal dynamics.
The integrated TWM achieves the highest SCM in every scenario (avg~45\%) and $\Delta_\mathrm{cost}{=}7\%$, supporting the value of integrating all three layers.

\begin{figure}[t]
\centering
\includegraphics[width=\columnwidth]{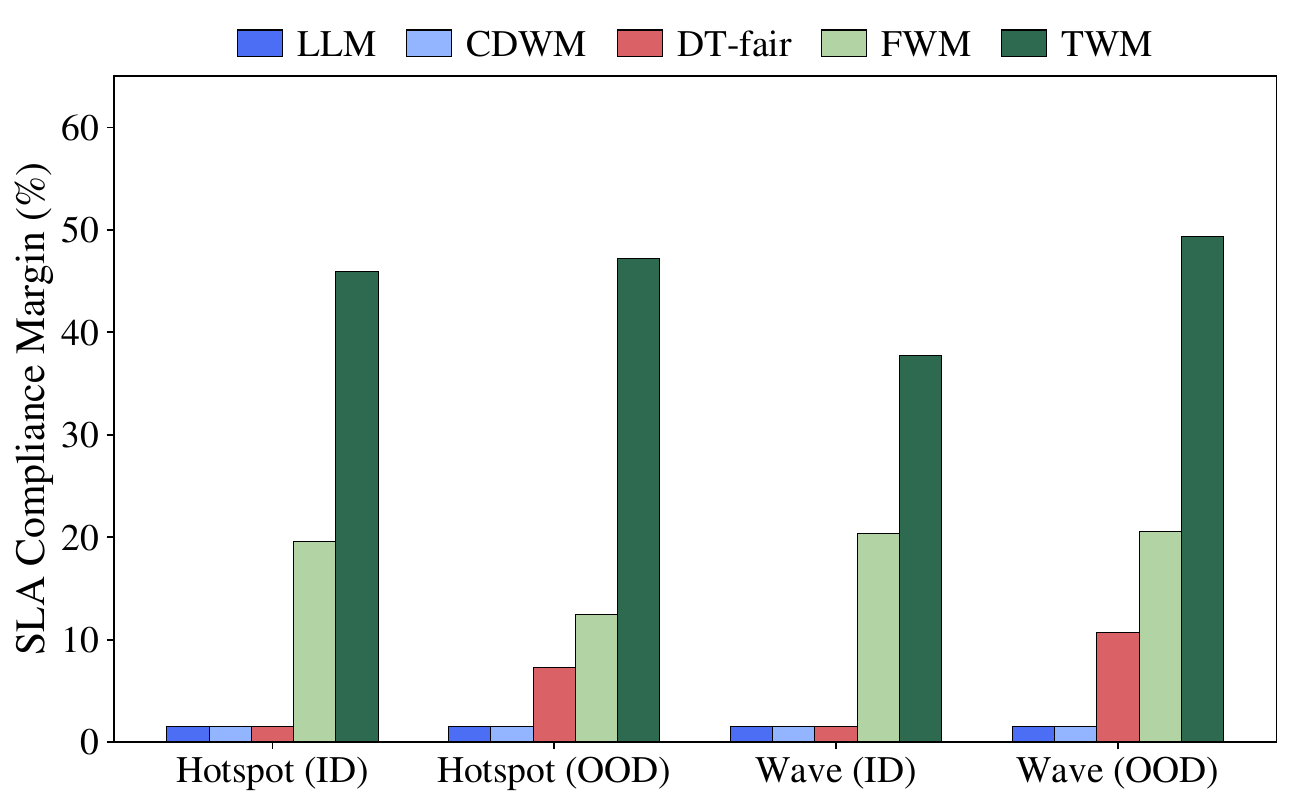}
\caption{SLA Compliance Margin by method and scenario (higher is better).  TWM achieves the highest margin (38--49\%) across all scenarios.  FWM achieves moderate margin but no cost savings.  DT-fair, CDWM, and LLM are near zero.}
\label{fig:ablation}
\end{figure}

Despite having the same search budget, DT-fair can only achieves avg $\mathrm{SCM}{=}4.5\%$.
The failure is not due to the lack of search effort, but the tendency of exhaustive search over a fixed candidate library to select cost-efficient policies that sacrifice individual slices. In particular, FWA loss reaches 24.7\% in ID scenarios and eMBB loss exceeds 25\% in OOD scenarios. OOD robustness is stable: TWM's SCM ranges from 38\% (wave-ID) to 49\% (wave-OOD) with no systematic degradation, whereas DT-fair remains fragile under distribution shift.
These results provide validation that integrating field, dynamics, and language-model layers yields planning outcomes that single-layered baselines cannot achieve. Improving autonomous policy selection in the CDWM and scaling to larger multi-site settings is important step in our future work.

\section{Challenges and Future Directions}
\label{sec:challenges}

While the TWM architecture is now well-defined, realizing production-grade systems introduces substantial open challenges. In the following we discuss four key areas.

\textbf{Theory and Foundations}
The TWM’s connection to POMDPs and hierarchical state-space models raises several theoretical questions. Identifiability concerns the conditions under which the FWM and CDWM can jointly recover the latent telecom state from partial and noisy observations. Causal structure relates to how the CDWM can distinguish the effect of actions from correlations induced by the logging policy, particularly for off-policy evaluation of configurations that were never deployed. Uncertainty quantification addresses how epistemic and aleatoric uncertainty can be propagated through the multi-layer model and translated into calibrated confidence bounds on KPI predictions, enabling downstream planning with reliable guarantees.


\textbf{Algorithms and Training}
Training a multi-layer TWM raises challenges beyond those of standard world models. The FWM and CDWM operate at different timescales and data modalities, hence, joint or hierarchical training schemes that avoid catastrophic forgetting across layers are needed.  Physics-informed priors, such as 3GPP channel models or queuing-theoretic approximations, can regularize learning and improve sample efficiency, but integrating them into modern deep-learning pipelines without sacrificing flexibility remains an open problem.  Continual and online adaptation is essential for tracking non-stationary network conditions, yet must be paired with safety constraints that prevent model drift from degrading operational decisions.

\textbf{Systems and Deployment}
Deploying a TWM in a live network requires the integration with existing infrastructure, including O-RAN RIC, NDTs, OSS/BSS platforms, and telemetry pipelines. Practical questions include where to host each layer (edge, regional cloud, or central cloud), how to manage the latency--accuracy trade-off between lightweight local models and more capable but remote ones, and how to version, validate, and roll back models in a continuous-deployment setting. A related challenge is the development of standardized benchmarks and evaluation protocols, a {TelecomWorldBench}, that can measure prediction accuracy, decision quality, and robustness to distribution shifts in a reproducible and comparable manner.

\textbf{Safety, Governance, and Trust}
Autonomous network management is safety-critical, a misconfigured action can cause SLA violations, revenue loss, or cascading failures. Building operator trust requires not only accurate predictions but also transparent explanations that trace each recommendation to its world-model evidence, auditable logs of every planning step, and progressive autonomy frameworks in which the TWM earns increasing decision authority as its track record accumulates.  Regulatory and data-governance considerations, including privacy-preserving training, compliance with spectrum and service obligations, and clear accountability for model-driven decisions, must be addressed as TWMs move from research prototypes toward operational deployment.

\section{Conclusion}
\label{sec:conclusion}
In this articule, we introduced the Telecom World Model (TWM), a three-layer architecture that separates network dynamics into a field world model (FWM) for spatial prediction, a control/dynamics world model (CDWM) for action-conditioned KPI forecasting, and a telecom foundation model for intent-driven planning. The design is structured around a two-world perspective that distinguishes controllable network state from exogenous environment. This formulation addresses a gap that is not captured by existing approaches such as LLM-based agents, digital twins, or standalone world models. In particular, we highlight conditional diffusion models and latent dynamics models as practical candidates for the FWM and CDWM layers. A proof-of-concept on multi-domain network slicing shows that the full TWM can achieve SLA compliance while reducing cost, compared to both single-paradigm baselines and digital twin-based search under similar resource constraints. Instead of replacing LLMs or digital twins, TWM provides a framework in which they can be combined, with the world model acting as the predictive core that grounds higher-level reasoning in network dynamics.


\bibliographystyle{IEEEtran}
\bibliography{ref} 

\end{document}